\documentclass[11pt]{article}

\usepackage{acl}

\usepackage{enumitem}
\usepackage{times}
\usepackage{latexsym}
\usepackage[T1]{fontenc}
\usepackage[utf8]{inputenc}
\usepackage{microtype}
\usepackage{inconsolata}
\usepackage{graphicx}
\usepackage{booktabs}
\usepackage{amsmath}
\usepackage{amssymb}
\usepackage{amsfonts}
\usepackage{multirow}
\usepackage{url}
\usepackage[table,dvipsnames]{xcolor}
\usepackage{pifont}
\usepackage[most]{tcolorbox}
\usepackage{xspace}

\definecolor{my_green}{RGB}{40,154,121}
\definecolor{my_red}{RGB}{176,46,46}
\definecolor{backblue}{RGB}{244,247,251}
\definecolor{LastBule}{RGB}{192,211,235}
\definecolor{LastYellow}{RGB}{244,245,208}
\definecolor{myblue}{RGB}{39,108,191}
\definecolor{mypink}{RGB}{244,231,250}
\definecolor{myorange}{RGB}{255,236,231}
\definecolor{bggreen}{RGB}{224,236,233}
\definecolor{mypurple}{RGB}{217,205,246}

\newcommand{\correctmark}{\textcolor{my_green}{\ding{52}}}
\newcommand{\errormark}{\textcolor{my_red}{\ding{56}}}
\newcommand{\best}[1]{\cellcolor{NavyBlue!30}\textbf{#1}}
\newcommand{\second}[1]{\cellcolor{NavyBlue!10}\underline{#1}}
\newcommand{\eg}{\hbox{\emph{e.g.},}\xspace}
\newcommand{\ie}{\hbox{\emph{i.e.},}\xspace}
\newcommand{\homepage}{\raisebox{-1.5pt}{\includegraphics[height=1.2em]{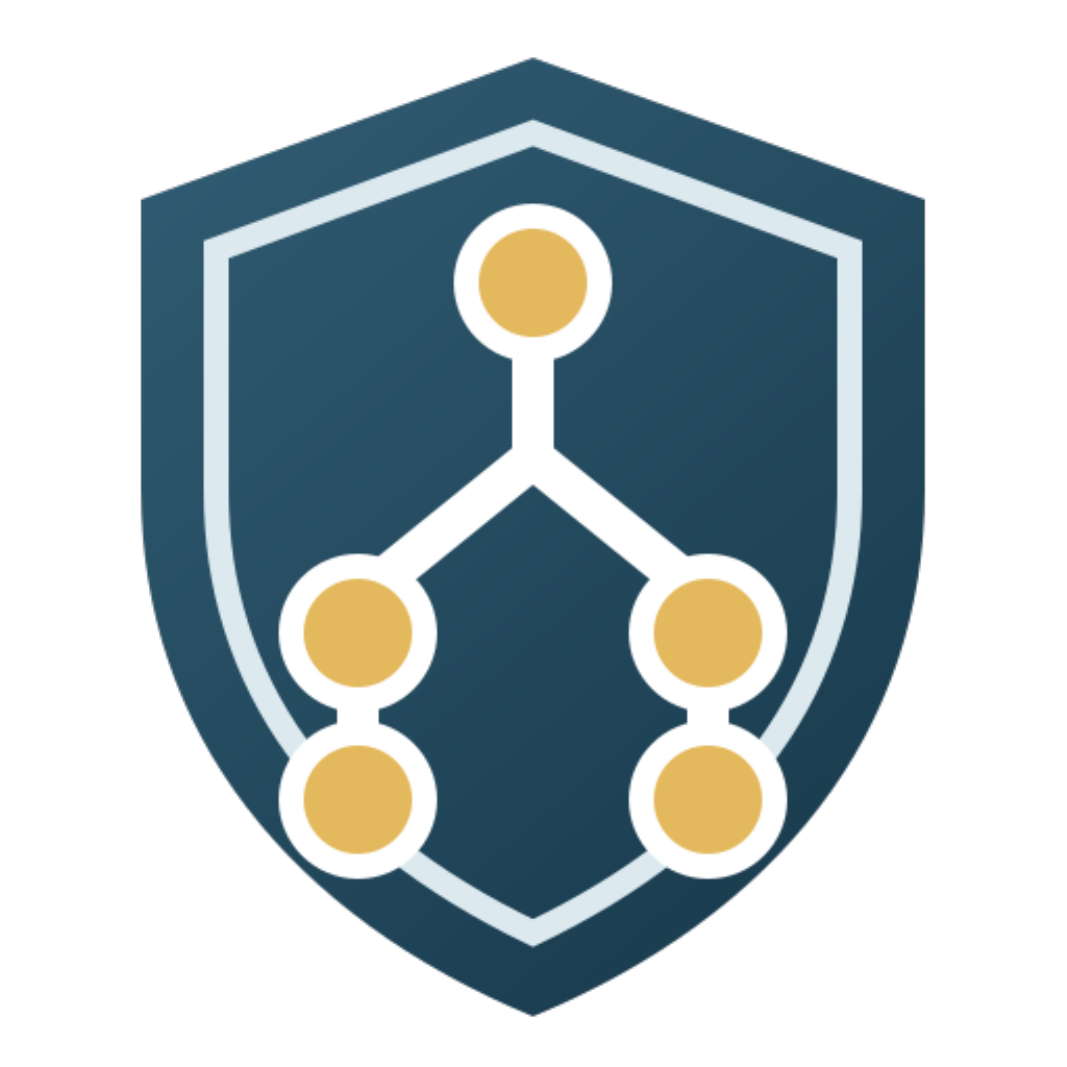}}}
\newcommand{\github}{\raisebox{-1.5pt}{\includegraphics[height=1em]{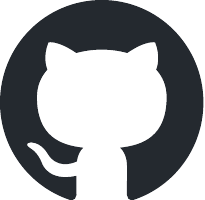}}}
\newcommand{\huggingface}{\raisebox{-1.5pt}{\includegraphics[height=1em]{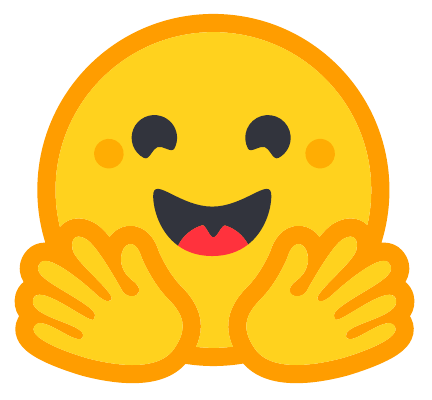}}}

\title{Can LLMs Follow Medical Expert Logic? A Benchmark for Hierarchical Logical Consistency in Risk-of-Bias Assessment}

\author{
  \textbf{Jiayu Huang}\textsuperscript{\rm 1} \quad
  \textbf{Zichen Tang}\textsuperscript{\rm 1} \quad
  \textbf{Qianhui Ling}\textsuperscript{\rm 2} \quad
  \textbf{Zemin Kuang}\textsuperscript{\rm 2}\thanks{Corresponding authors.} \quad
  \textbf{Haihong E}\textsuperscript{\rm 1*}\\
  \textsuperscript{\rm 1}Beijing University of Posts and Telecommunications\\
  \textsuperscript{\rm 2}Hypertension Center, Beijing Anzhen Hospital, Capital Medical University\\
  \homepage~~\href{https://bupt-reasoning-lab.github.io/LogiMed-RoB}{\texttt{bupt-reasoning-lab.github.io/LogiMed-RoB}}\\
  \github~~\href{https://github.com/BUPT-Reasoning-Lab/LogiMed-RoB}{\texttt{BUPT-Reasoning-Lab/LogiMed-RoB}}
  \quad
  \huggingface~~\href{https://huggingface.co/datasets/BUPT-Reasoning-Lab/LogiMed-RoB}{\texttt{BUPT-Reasoning-Lab/LogiMed-RoB}}
}

\begin{document}
\maketitle

\begin{abstract}
Evidence-based medicine demands strict logical consistency, yet current evaluations of large language models (LLMs) prioritize superficial label matching over genuine reasoning. We introduce \textbf{LogiMed-RoB}, a benchmark grounded in Cochrane Risk of Bias (RoB) 2.0 expert logic, comprising 860 randomized controlled trials (RCTs) and 14,820 queries. It evaluates models under the \textbf{Hierarchical Logical Consistency (HLC)} framework across four dimensions: Atomic Consistency, Domain Consistency, Aggregation Consistency, and Evidential Faithfulness.
Experiments on 10 state-of-the-art LLMs reveal a catastrophic \textit{Error Compounding Effect}: despite the top model reaching $98.88\%$ Atomic Consistency, its end-to-end consistency collapses to $45.13\%$, with several open-weight architectures plummeting to nearly $0\%$. We further uncover a systematic \textit{evidence-reasoning gap}: even when models retrieve high-quality evidence, they fail to deduce correct outcomes in 18.63--40.05\% of cases, while Blind Guess Rates reach $48.28\%$. LogiMed-RoB demonstrates that high outcome accuracy can conceal critical reasoning flaws, underscoring the necessity of white-box logical verification for clinical deployment.
\end{abstract}

\section{Introduction}

With large language models (LLMs) achieving saturation on standard medical benchmarks~\citep{jin2021disease, singhal2023large, nori2023capabilities}, the field is shifting toward expert-level benchmarks that emphasize clinical depth.
Datasets such as MedXpertQA~\citep{zuo2025medxpertqa} aim to stress-test models in clinical scenarios, moving evaluation from simple answer accuracy toward clinical compliance.
However, most current datasets remain in a \textit{result-oriented} paradigm, focusing on superficial alignment between model outputs and gold labels~\citep{wang2025scores, agrawal2024evaluation}.
This focus overlooks the rigorous nature of medical reasoning and allows models to reach the correct result through flawed logic. Consequently, it creates an illusion of high performance while hiding real clinical risks~\citep{gu2025illusion}.

\begin{figure}[t]
  \centering
  \includegraphics[width=\columnwidth]{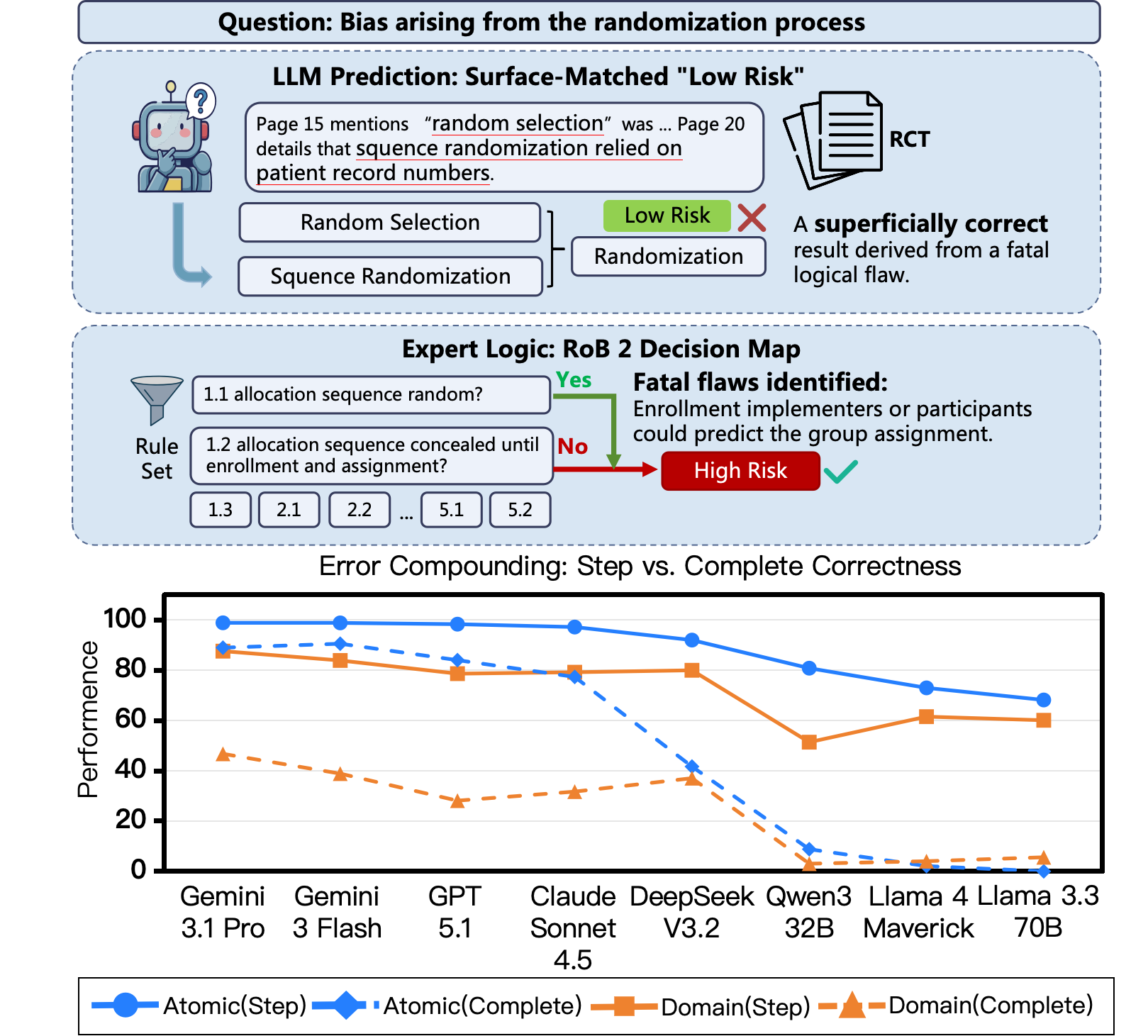}
  \caption{The figure illustrates the reasoning gap in risk-of-bias assessment. While LLMs often make incorrect predictions through superficial keyword matching, the rigorous RoB~2.0 decision map reveals a severe \textit{Error Compounding Effect}: atomic-level logical deviations propagate through multi-step synthesis, causing erroneous end-to-end judgments.}
  \label{fig:question_sample}
\end{figure}

\begin{table*}[tbp]
\centering
\footnotesize
\resizebox{0.97\textwidth}{!}{%
\addtolength{\tabcolsep}{-0.3em}
\renewcommand{\arraystretch}{1.1}
\begin{tabular}{l c c c c c c c c c}
\toprule
\multirow{2}{*}{\textbf{Method}} &
\multirow{2}{*}{\textbf{RCT Samples}} &
\multirow{2}{*}{\textbf{RoB Version}} &
\multirow{2}{*}{\textbf{Question Level}} &
\multicolumn{4}{c}{\textbf{Capability Coverage}} &
\multirow{2}{*}{\textbf{Year}} &
\multirow{2}{*}{\textbf{Detection Type}} \\
\cmidrule(lr){5-8}
& & & & Retrieval & Cond.~Reasoning & Expert Align. & Factual Adher.~\& Attrib. & & \\
\midrule
RobotReviewer$^\dagger$  & 12,808 & RoB~1.0 & Domain  & \correctmark & \errormark & \errormark & \errormark & 2016 & SVM \\
ROBIN                    & 4,562  & RoB~1.0 & Domain  & \correctmark & \errormark & \errormark & \correctmark & 2024 & LLM \\
RoBBR                    & 700    & RoB~1.0 & Domain  & \correctmark & \errormark & \errormark & \correctmark & 2024 & LLM \\
URSE$^\dagger$           & 467    & RoB~1.0 & Domain  & \correctmark & \errormark & \errormark & \errormark & 2024 & SVM \\
LLMPatchbay$^\dagger$    & 100    & RoB~2.0 & Domain  & \errormark & \errormark & \errormark & \errormark & 2024 & LLM \\
ROBOTO2$^\dagger$        & 521    & RoB~2.0 & Domain  & \errormark & \errormark & \errormark & \errormark & 2025 & LLM \\
RoB-Item$^\dagger$       & 53     & RoB~2.0 & Atomic  & \errormark & \errormark & \errormark & \errormark & 2025 & LLM \\
RoB-Domain$^\dagger$     & 319    & RoB~2.0 & Domain  & \errormark & \errormark & \errormark & \errormark & 2025 & LLM \\
GEPA$^\dagger$           & 100    & RoB~1.0 & Domain  & \errormark & \errormark & \errormark & \errormark & 2025 & LLM \\
\midrule
\textbf{LogiMed-RoB (ours)} & \textbf{860} & \textbf{RoB~1.0/2.0} & \textbf{Atomic} & \textbf{\correctmark} & \textbf{\correctmark} & \textbf{\correctmark} & \textbf{\correctmark} & \textbf{2026} & \textbf{LLM} \\
\bottomrule
\end{tabular}
}
\caption{Comparison with prior RoB benchmarks. \textbf{Cond.~Reasoning}: Conditional reasoning; \textbf{Expert Align.}: Expert rule alignment; \textbf{Factual Adher.~\& Attrib.}: Factual adherence \& attribution. $^\dagger$ indicates that the dataset is not publicly released.}
\label{tab:comparison}
\vspace{-0.3cm}
\end{table*}

Evidence-based medicine (EBM) emphasizes the logical integrity of the entire chain from clinical question to final decision~\citep{Sackett71}. As the core methodology for evaluating the reliability of medical research findings, Risk-of-Bias (RoB) assessment is used to identify systematic errors in clinical trials that may cause results to deviate from the truth~\citep{cochrane2022handbook,higgins2019rob}. RoB assessment is inherently a hierarchical reasoning process rather than a classification performed in a single step. As illustrated in Figure~\ref{fig:question_sample}, this process ranges from extracting granular details to deducing domain risks and synthesizing a global decision. However, current benchmarks predominantly evaluate end-to-end label accuracy. If a model achieves the correct final label via spurious correlations while violating established medical logic, it fundamentally fails the strict requirements of clinical trustworthiness~\citep{ghassemi2021false}.

To bridge this gap, we propose \textbf{Hierarchical Logical Consistency (HLC)}, a post-hoc audit framework.
We formalize the Cochrane RoB 2.0 gold standard as deterministic expert rules and systematically verify whether LLM outputs conform to the established clinical decision process.
We construct \textbf{LogiMed-RoB}, comprising 860 randomized controlled trials (RCTs) and 14,820 queries, to evaluate model capabilities along four dimensions: (1) \textbf{Atomic Consistency} quantifies the rigor of conditional reasoning at the base node level;
(2) \textbf{Domain Consistency} measures the alignment between model output and expert rule derivation;
(3) \textbf{Aggregation Consistency} verifies compliance with the macroscopic ``Worst-of Principle'';
and (4) \textbf{Evidential Faithfulness} assesses whether decisions are anchored in accurate textual evidence rather than driven by parametric priors.

In our study, we conduct extensive experiments using LogiMed-RoB to evaluate 10 representative LLMs. We observe a catastrophic \textit{Error Compounding Effect}: while Gemini 3.1 Pro achieves near-perfect Atomic Consistency (98.88\%), its end-to-end consistency collapses to 45.13\%, with several open-weight models plummeting to $\sim$0\%. We also discover a systematic \textit{evidence-reasoning gap}. Even when models successfully retrieve high-quality evidence, they persistently fail to deduce correct outcomes, with Reasoning Failure Rates ranging from 18.63\% to 40.05\%, while Blind Guess Rates reach 48.28\%. By evaluating models under HLC, we show how high outcome accuracy can conceal structural fragilities, demonstrating the necessity of white-box logical verification in high-stakes EBM.

\section{Related Work}

\begin{figure*}[t]
  \centering
  \includegraphics[width=0.97\textwidth]{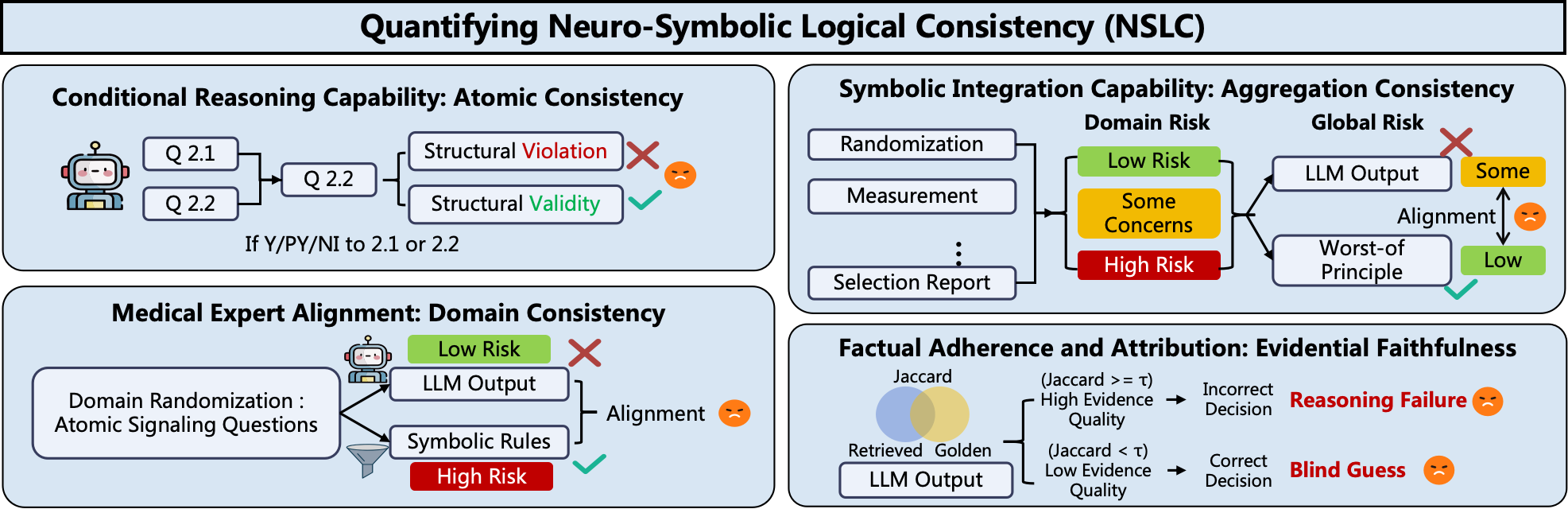}
  \caption{The Hierarchical Logical Consistency (HLC) framework formalizes RoB guidelines as deterministic expert rules and audits LLM outputs across four dimensions: Atomic Consistency, Domain Consistency, Aggregation Consistency, and Evidential Faithfulness.}
  \label{fig:HLC}
\end{figure*}

\subsection{Automation of Risk-of-Bias Assessment}

RoB~1.0~\citep{Higginsd5928} organized assessments around broad bias domains and required assessors to make a direct risk judgment within each domain. RobotReviewer~\citep{marshall2016robotreviewer} automated several of these domain-level judgments while identifying supporting text. EBM-NLP~\citep{nye2018corpus}, by contrast, provided PICO annotations for evidence extraction rather than RoB labels. Neither resource was designed to audit whether intermediate outputs obey a multilevel decision hierarchy.

RoB~2.0~\citep{higgins2019rob} introduced signaling questions and algorithms that map their responses to proposed domain-level judgments. The overall judgment is then synthesized from the domain judgments and generally reflects the most severe domain, with specified escalation when several domains raise concerns. Modern automated assessments must therefore advance from label matching to faithfully executing this multistep decision process.

However, as summarized in Table~\ref{tab:comparison}, existing benchmarks exhibit three critical limitations:
\begin{itemize}[leftmargin=*]
    \item \textbf{Coarse Granularity}\textnormal{:} Several datasets, including RoBIn~\citep{dias2025robin}, the main task in RoBBR~\citep{bergen2024measuring}, and GEPA~\citep{li2025gepa}, focus on domain-level risk labels rather than cross-level consistency among signaling-question responses, domain judgments, and overall judgments.
    \item \textbf{Cross-Level Evaluation Gap}\textnormal{:} RoB~2.0 resources such as LLMPatchbay~\citep{EiseleMetzgeretal2025}, ROBOTO2~\citep{hevia2025roboto2}, and RoB-Domain~\citep{robguard2025} support domain judgments or signaling-question assessment. However, they do not explicitly audit consistency among atomic answers, rule-derived domain labels, and direct domain judgments. This omission can still reward shortcut learning~\citep{geirhos2020shortcut}.
    \item \textbf{Accessibility}\textnormal{:} Several datasets (marked $^\dagger$) were not released as standalone public resources by their source papers, which limits reproducible comparison.
\end{itemize}

\textbf{LogiMed-RoB} directly addresses this gap. While RoB-Item~\citep{robguard2025} operates at the atomic level, it evaluates isolated signaling questions without auditing cross-level dependencies. To our knowledge, LogiMed-RoB is the first publicly available benchmark to jointly combine atomic-level verification, hierarchical conditional-reasoning audits, and end-to-end consistency tracking across all three levels. It systematically evaluates the four logical consistency dimensions essential for clinical trustworthiness.

\subsection{LLM Logical Reasoning and Rule-Based Auditing}

LLMs remain fragile on multistep logical tasks~\citep{marcus2020next, huang2023towards} and often fail to generalize compositionally~\citep{dziri2023faith}. Chain-of-thought prompting~\citep{wei2022chain} can also produce explanations that do not faithfully reflect the factors driving a prediction~\citep{turpin2023language}. Logic-LM~\citep{pan2023logiclm} and LINC~\citep{olausson2023linc} combine language models with external symbolic solvers, while self-consistency~\citep{wang2022self} aggregates answers across multiple sampled reasoning paths. These methods address general logical reasoning rather than auditing compliance with a domain-specific hierarchy that links signaling questions, domain judgments, and global aggregation.

We address this with the \textbf{Hierarchical Logical Consistency (HLC)} framework. HLC formalizes the Cochrane RoB~2.0 decision system as a deterministic rule set $\mathcal{R}$ and audits whether LLM outputs conform to the expert decision hierarchy. This post-hoc approach quantifies logical alignment across Atomic Consistency, Domain Consistency, Aggregation Consistency, and Evidential Faithfulness, enabling fine-grained diagnosis of \textit{where} and \textit{how} models fail in structured clinical reasoning.

\section{Benchmark}

\subsection{A Primer on RoB Workflows}

\paragraph{The RoB 2.0 Logical Workflow\textnormal{:}}
As illustrated in our Hierarchical Logical Consistency (HLC) framework (Figure~\ref{fig:HLC}), RoB 2.0 operates as a strict, algorithmic pipeline that eliminates subjective guesswork through three sequential steps:
(1) \textbf{Atomic Fact-Checking}\textnormal{:} Experts answer highly specific, factual ``Signaling Questions'' regarding trial methodology (\eg patient blinding).
(2) \textbf{Domain-Level Deduction}\textnormal{:} A predefined clinical ``decision map'' (detailed in Appendix~\ref{sec:app_rob_rules}) automatically routes these atomic answers to a risk level (\textit{Low}, \textit{Some Concerns}, or \textit{High}) for specific bias domains.
(3) \textbf{Global Aggregation}\textnormal{:} The overall risk of the trial is determined via a strict ``Worst-of Principle'' (\eg a single \textit{High} risk domain flags the entire trial as \textit{High} risk).

\paragraph{The RoB 1.0 Evidence-Justification Workflow\textnormal{:}}
Beyond internal reasoning, evaluating textual grounding is fundamental to clinical trustworthiness. The RoB 1.0 standard explicitly requires experts to extract exact sentences justifying each risk judgment (detailed in Appendix~\ref{sec:app_rob_rules_rob1}). By adopting these human-annotated quotes as the gold standard, we audit a model's \textbf{Evidential Faithfulness}. This assessment verifies whether a predicted risk label is grounded in the correct underlying evidence or is merely a lucky ``Blind Guess'', ensuring that LLM conclusions stem from factual text rather than spurious correlations.

\subsection{Problem Formalization}

\paragraph{Notation and State Space\textnormal{:}} Let $\mathcal{D}_{test} = \{\mathcal{X}_1, \dots, \mathcal{X}_N\}$ denote $N$ RCT texts. The RoB 2.0 expert system is a tuple $\mathcal{S} = \langle \mathcal{Q}, \mathcal{A}, \mathcal{D}, \mathcal{L}, \mathcal{R} \rangle$, where $\mathcal{Q}$ is the set of signaling questions with answer space $\mathcal{A}=\{\text{Yes}, \text{PY}, \text{PN}, \text{No}, \text{NI}, \text{NA}\}$, $\mathcal{D}$ denotes five bias domains, $\mathcal{L}=\{\text{Low}, \text{Some Concerns}, \text{High}\}$ is the set of ranked risk labels ($\text{Low} < \text{Some Concerns} < \text{High}$), and $\mathcal{R}$ is the deterministic rule set.

\paragraph{Level 0 Neural Fact Generation\textnormal{:}} Model $f_{\theta}$ extracts a discrete judgment $\hat{a}_{n,i,j} \in \mathcal{A}$ for each signaling question: $\hat{a}_{n,i,j} = f_{\theta}(\mathcal{X}_n, q_{i,j})$. Together, these judgments form the \textit{Neural Fact Base}.

\paragraph{Level 1 Atomic Fact-Checking\textnormal{:}} In a directed acyclic graph, node validity $v_{n,i,j} \in \{0, 1\}$ depends on parent answers via $v_{n,i,j}=\mathcal{B}_{i,j}(\mathbf{A}_{n,Pa(q_{i,j})})$, defaulting to $1$ for root nodes. Atomic Consistency requires $\forall q_{i,j} \in d_i, \hat{a}_{n,i,j}=\text{NA} \Leftrightarrow v_{n,i,j}=0$. A violation of this condition indicates a failure to follow the clinical decision map.

\paragraph{Level 2 Domain-Level Deduction\textnormal{:}} The symbolic domain risk conclusion $y_{n,i}^{\text{symbolic}}$ is uniquely determined by activated answers using the Cochrane mapping operator, expressed as $y_{n,i}^{\text{symbolic}}=\mathcal{M}_i(\{\hat{a}_{n,i,j} \mid v_{n,i,j}=1\})$. Here, $y_{n,i}^{\text{symbolic}}$ represents the standard risk derived directly from professional clinical guidelines. If the model's direct neural output $\hat{y}_{n,i}^{\text{neural}}$ contradicts this deduction, the model fails to align with medical experts, indicating a critical breakdown in clinical trustworthiness.

\paragraph{Level 3 Global Aggregation\textnormal{:}} Global decisions follow the ``Worst-of Principle'': $Y_{n,global}^{\text{symbolic}} = \max_{d_i \in \mathcal{D}} \{y_{n,i}^{\text{symbolic}}\}$. A structural violation occurs if the model's predicted global risk strictly underestimates this bound (\ie $\hat{Y}_{n,global}^{\text{neural}} < \max_{d_i \in \mathcal{D}} \{y_{n,i}^{\text{symbolic}}\}$), breaking the predefined symbolic aggregation constraints. If the LLM generates a final conclusion that is inconsistent with this aggregation logic, it violates the clinical inference process used by human practitioners.

\subsection{Overview of LogiMed-RoB}

To systematically evaluate the Hierarchical Logical Consistency of LLMs in EBM, we constructed LogiMed-RoB.
LogiMed-RoB extends beyond a text classification dataset by providing an expert-level evaluation framework with multilevel reasoning chains.

\subsubsection{Dual-Track Evaluation Paradigm}

Reflecting the structural divergence between RoB standards, we establish a Dual-Track Evaluation Paradigm comprising 860 randomized controlled trials (Table~\ref{tab:dataset}). To construct this benchmark, we sourced all papers exclusively from publicly available datasets and websites, ensuring that every document was available for public download. We also rigorously anonymized any content within the dataset that might contain personal information. This design isolates the hierarchical logic of RoB 2.0 from the textual grounding of RoB 1.0, thereby shifting the evaluation from flat label matching to a rigorous audit of multilevel reasoning and Evidential Faithfulness.

\begin{table}[tbp]
\centering
\small
\begin{tabular}{@{}lcc@{}}
\toprule
 & \textbf{Track A} & \textbf{Track B} \\ \midrule
\textbf{Standard} & Cochrane RoB 2.0 & Cochrane RoB 1.0 \\
\textbf{RCTs} & 201 & 659 \\
\textbf{Items} & 626 & 1,048 \\
\textbf{Questions} & 13,772 & 1,048 \\
\textbf{Domains} & 5 & 6 \\
\textbf{Source} & \begin{tabular}[c]{@{}c@{}}Cochrane (467)\\ Figshare (159)\end{tabular} & \begin{tabular}[c]{@{}c@{}}ROBIN (713)\\ RoBBR (335)\end{tabular} \\
\textbf{Gold Labels} & Expert & Expert \\ \bottomrule
\end{tabular}
\caption{Overview of the LogiMed-RoB dataset. The table summarizes the two evaluation tracks, which comprise 860 randomized controlled trials.}
\label{tab:dataset}
\end{table}

\paragraph{Track A Hierarchical Logic Alignment (RoB 2.0)\textnormal{:}}
This track uses the \textbf{hierarchical decision logic} of RoB 2.0 to isolate internal reasoning mechanisms. It contains 626 outcome-level instances sourced from the Cochrane Library\footnote{\url{https://www.cochranelibrary.com}} (467 items) and Figshare\footnote{\url{https://figshare.com}} (159 items). The 22 atomic signaling questions per assessment yield 13,772 fine-grained queries for auditing three forms of symbolic consistency: \textbf{Atomic Consistency}, \textbf{Domain Consistency}, and \textbf{Aggregation Consistency}. This setup quantifies the model's alignment with the complex rule-based expert decision system embodied in the RoB 2.0 framework.

\paragraph{Track B Evidential Faithfulness (RoB 1.0)\textnormal{:}}
Beyond internal logic, Track B evaluates textual grounding using 1,048 items from 659 RCTs sourced entirely from two existing benchmarks, ROBIN\footnote{\url{https://github.com/phdabel/robin}} (713) and RoBBR\footnote{\url{https://github.com/RoBBR-Benchmark/RoBBR}} (335), rather than from new annotations. Our contribution is methodological: a \textit{window cropping} technique that condenses full-length texts into focused context and $\tau$-threshold BGR/RFR metrics that decompose performance into evidential grounding and reasoning capability. This track verifies whether a model's risk label relies on correct evidence or merely reflects a lucky ``Blind Guess''.

\paragraph{Data Quality and Human Validation\textnormal{:}} \textit{Track A:} Our author team, which included licensed physicians, manually extracted all 626 instances from Cochrane systematic reviews (467) and peer-reviewed Figshare deposits (159). The team calibrated these instances against the original published conclusions. All source RoB assessments were validated by multiple expert reviewers during the original peer-review process. The gold-standard labels therefore reflect expert-level consensus rather than relying on \textit{de novo} annotation by a single team. \textit{Track B:} Medical annotators re-verified approximately 20\% of the samples (210 items), achieving 93.5\% agreement on window accuracy. No AI-generated pseudo-labels were used in either track.

\subsection{Hierarchical Logical Consistency Metrics}

\paragraph{Atomic Consistency\textnormal{:}} We define Atomic Consistency as the \textbf{Condition-Aware Ratio (CAR)}. Let $\mathcal{Q}_{cond}$ be the set of prerequisite-constrained questions.
\begin{equation}
CAR = \frac{1}{N |\mathcal{Q}_{cond}|} \sum_{n, q} \mathbb{I}(\hat{a}_{n,q} \!=\! \text{NA} \Leftrightarrow v_{n,q} \!=\! 0).
\end{equation}
It captures two valid states: correct pruning (outputting NA when precondition $v=0$) and correct activation (outputting a valid answer when $v=1$).

\paragraph{Domain Consistency\textnormal{:}} We define \textbf{Logical Fidelity (LF)} to evaluate structural integrity across the five standard domains of the RoB 2.0 framework. It measures the match between the model's direct neural output $\hat{y}_{n,i}^{\text{neural}}$ and the symbolic deduction label $y_{n,i}^{\text{symbolic}}$ derived from its atomic answers within each domain.

\begin{equation}
LF = \frac{1}{5N} \sum_{n=1}^{N} \sum_{i=1}^{5} \mathbb{I}\!\left( \hat{y}_{n,i}^{\text{neural}} = y_{n,i}^{\text{symbolic}} \right).
\end{equation}

\paragraph{Aggregation Consistency\textnormal{:}} We use \textbf{Verification Rate (VR)} to evaluate whether the model adheres to the ``Worst-of Principle'':
\begin{equation}
VR = \frac{1}{N} \sum_{n=1}^N \mathbb{I}\!\left( \hat{Y}_{n,global}^{\text{neural}} = Y_{n,global}^{\text{symbolic}} \right),
\end{equation}
where $\hat{Y}_{n, \text{global}}^{\text{neural}}$ denotes the global risk level predicted directly by the model's neural inference, and $Y_{n, \text{global}}^{\text{symbolic}}$ represents the ground-truth label strictly derived from the valid domain-level prediction set according to the symbolic rule set $\mathcal{R}$.

\paragraph{Evidential Faithfulness\textnormal{:}} We use Jaccard similarity to quantify \textbf{Retrieval Strength}.
To distinguish genuine reasoning from spurious correlation, we introduce a predefined threshold $\tau$ to separate strong from weak evidence matching. We then define the \textbf{Blind Guess Rate (BGR)} and the \textbf{Reasoning Failure Rate (RFR)}:
\begin{equation}
\begin{aligned}
\text{BGR} &= P(\text{Correct} \wedge \text{Jaccard} < \tau) \\
\text{RFR} &= P(\text{Incorrect} \mid \text{Jaccard} \geq \tau).
\end{aligned}
\end{equation}
BGR measures the absolute prevalence of ungrounded correct predictions across the entire evaluation set, avoiding the denominator effect. RFR retains a conditional form, as it specifically isolates reasoning failures within the well-retrieved subpopulation.

\begin{table*}[t]
\centering
\small
\setlength{\tabcolsep}{3.0pt}
\begin{tabular}{l cccc cccccc c cc} 
\toprule
\textbf{Model} & \multicolumn{4}{c}{\textbf{Atomic (CAR)}} & \multicolumn{6}{c}{\textbf{Domain (LF)}} & \textbf{VR} & \multicolumn{2}{c}{\textbf{Evidence}} \\
\cmidrule(lr){2-5} \cmidrule(lr){6-11} \cmidrule(lr){13-14}
& D2 & D3 & D4 & Overall & D1 & D2 & D3 & D4 & D5 & Overall & & BGR & RFR \\
\midrule
\multicolumn{14}{l}{\textit{Proprietary}} \\
GPT 5.1 & 98.16 & 97.02 & \second{99.89} & 98.34 & 80.67 & 48.24 & 81.15 & 95.84 & 87.38 & 78.65 & 96.17 & 36.64 & 26.68 \\
Claude Sonnet 4.5 & 98.18 & \best{97.46} & 95.63 & 97.20 & 88.67 & 55.34 & \second{88.03} & 85.90 & 78.16 & 79.22 & \best{100.00} & 35.78 & 25.87 \\
Gemini 3 Flash & \second{99.88} & \second{97.39} & 98.93 & \second{98.85} & \second{92.96} & 50.24 & \best{90.56} & \second{96.96} & \second{88.80} & \second{83.90} & 94.24 & 46.28 & 23.24 \\
Gemini 3.1 Pro & \best{99.91} & 96.39 & \best{100.00} & \best{98.88} & \best{95.13} & \second{58.48} & 87.91 & \best{97.47} & \best{99.10} & \best{87.61} & \second{99.64} & \second{35.50} & \best{18.63} \\
\midrule
\multicolumn{14}{l}{\textit{Open-weight}} \\
DeepSeek V3.2 & 95.57 & 94.46 & 84.82 & 92.01 & 91.85 & \best{71.73} & 72.52 & 76.48 & 87.38 & 79.99 & \best{100.00} & 36.74 & 26.15 \\
Llama 3.3 70B & 71.01 & 47.60 & 84.98 & 68.18 & 90.87 & 22.72 & 45.76 & 63.84 & 77.32 & 60.10 & \best{100.00} & 39.03 & 27.15 \\
Qwen3 32B & 88.60 & 74.56 & 76.75 & 80.83 & 88.14 & 49.28 & 40.96 & 30.00 & 48.48 & 51.39 & 99.36 & \second{32.44} & 28.32 \\
Llama 4 Maverick & 76.69 & 49.35 & 91.72 & 73.00 & 92.08 & 47.33 & 19.03 & 72.37 & 76.94 & 61.54 & \second{99.84} & 48.28 & \second{22.42} \\
Baichuan M2 & 85.36 & 80.32 & 92.05 & 85.86 & 82.88 & 41.60 & 49.92 & 64.74 & 45.44 & 56.91 & 99.84 & \best{31.58} & 40.05 \\
Baichuan M3 & 87.68 & 97.37 & 99.12 & 94.02 & 89.82 & 35.26 & 84.39 & 89.46 & 63.86 & 72.55 & 99.30 & 33.97 & 25.06 \\
\bottomrule
\end{tabular}
\caption{Main Performance on HLC Metrics.
Atomic (CAR): Condition-Aware Ratio per domain (D2 to D4) and overall.
Domain (LF): Logical Fidelity per RoB 2.0 domain (D1 to D5) and overall.
VR: Verification Rate; BGR: Blind Guess Rate; RFR: Reasoning Failure Rate.}
\label{tab:main_results}
\end{table*}

\section{Experiments}

\subsection{Models and Experimental Setup}

We evaluate 10 LLMs: four proprietary models (GPT 5.1~\citep{openai2025gpt51}, Claude Sonnet 4.5~\citep{claude-sonnet-4.5}, Gemini 3 Flash~\citep{google2025gemini3flash}, and Gemini 3.1 Pro~\citep{google2026gemini31pro}) and six open-weight architectures (DeepSeek V3.2~\citep{deepseekai2025deepseekv32}, Llama 3.3 70B~\citep{meta2024llama33}, Qwen3 32B~\citep{yang2025qwen3technicalreport}, Llama 4 Maverick~\citep{meta2025llama4}, Baichuan M2~\citep{baichuanm2}, and Baichuan M3~\citep{baichuanm3}). To ensure reproducibility and eliminate stochastic noise, all models are accessed through official APIs with the temperature set to $T=0.0$. To assess Evidential Faithfulness, we set the Retrieval Strength threshold to $\tau=0.8$. Identical prompt templates are applied to all models, as provided in Appendix~\ref{app:prompt_templates}.

\subsection{Main Results}

Table~\ref{tab:main_results} presents a comprehensive evaluation of the 10 LLMs across the HLC metrics. The results reveal a pronounced dichotomy between the models' proficiency in rule-based branching and their severe limitations in multi-step logical synthesis. A systematic gap between evidential grounding and decision accuracy also emerges as a pervasive flaw across both open-weight and proprietary architectures. Appendix~\ref{sec:case_studies} provides detailed qualitative analyses and representative failure cases.

\paragraph{Atomic Consistency\textnormal{:}} The majority of proprietary models achieve near-perfect CAR ($>$97\%), with Gemini 3.1 Pro reaching a state-of-the-art CAR of 98.88\%.
While DeepSeek V3.2 demonstrates strong open-weight performance (92.01\%), models such as Llama 3.3 70B (68.18\%) and Llama 4 Maverick (73.00\%) exhibit significant structural weaknesses in handling conditional constraints, indicating that processing complex topological branching remains a challenge for certain architectures.

\begin{figure}[t]
  \centering
  \includegraphics[width=\columnwidth]{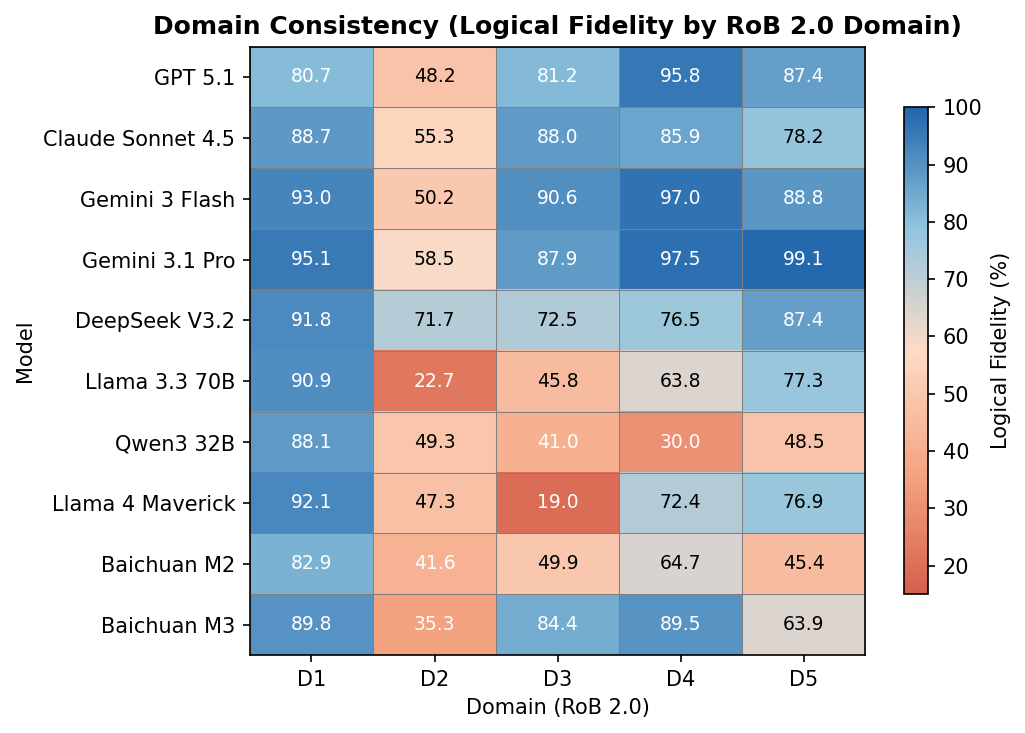}
  \caption{The heatmap shows Logical Fidelity by RoB~2.0 domain across all 10 evaluated models, with D2 and D3 emerging as critical consistency bottlenecks.}
  \label{fig:domain_consistency}
\end{figure}

\paragraph{Domain Consistency\textnormal{:}} Across all evaluated models, Logical Fidelity (LF) is consistently and significantly lower than the corresponding atomic performance.
Even the top-performing Gemini 3.1 Pro (87.61\%) and the leading open-weight model DeepSeek V3.2 (79.99\%) display a substantial performance gap.
As shown in Figure~\ref{fig:domain_consistency}, logical misalignment is not evenly distributed across the reasoning chain.
Rather, it is concentrated in specific domains, especially D2, which emerges as a critical consistency bottleneck across models.
This uneven degradation indicates that aggregating discrete local signals into domain-level conclusions remains the primary logical bottleneck.

\paragraph{Aggregation Consistency\textnormal{:}} Verification Rate (VR) exceeds 94\% across most models, with Claude Sonnet 4.5 and Llama 3.3 70B at 100.00\%, indicating that LLMs can effectively follow simple rules such as the Worst-of Principle.

\paragraph{Evidential Anchoring Deficit\textnormal{:}} Under a strict retrieval threshold ($\tau=0.8$), the Blind Guess Rate reveals that a substantial fraction of all predictions are ungrounded correct guesses, ranging from 31.58\% (Baichuan M2) to 48.28\% (Llama 4 Maverick).
This implies that in a majority of instances where models lack sufficient contextual evidence, they still output the ``correct'' label.
This spurious correctness confirms that models frequently bypass evidence-based deduction, relying heavily on parametric priors learned during pretraining.

\paragraph{Persistent Reasoning Failure\textnormal{:}} Conversely, the Reasoning Failure Rate demonstrates that even when high-quality evidence is successfully retrieved, models consistently fail to deduce the correct label.
Gemini 3.1 Pro records a lower rate of 18.63\%, while most models fall between 22\% and 28\%.
This ``evidence-reasoning gap'' demonstrates that current architectures can perform surface-level information extraction but consistently struggle to translate those extracted facts into robust logical derivations.

\subsubsection{Threshold Sensitivity Analysis}

\begin{figure*}[t]
  \centering
  \includegraphics[width=\textwidth]{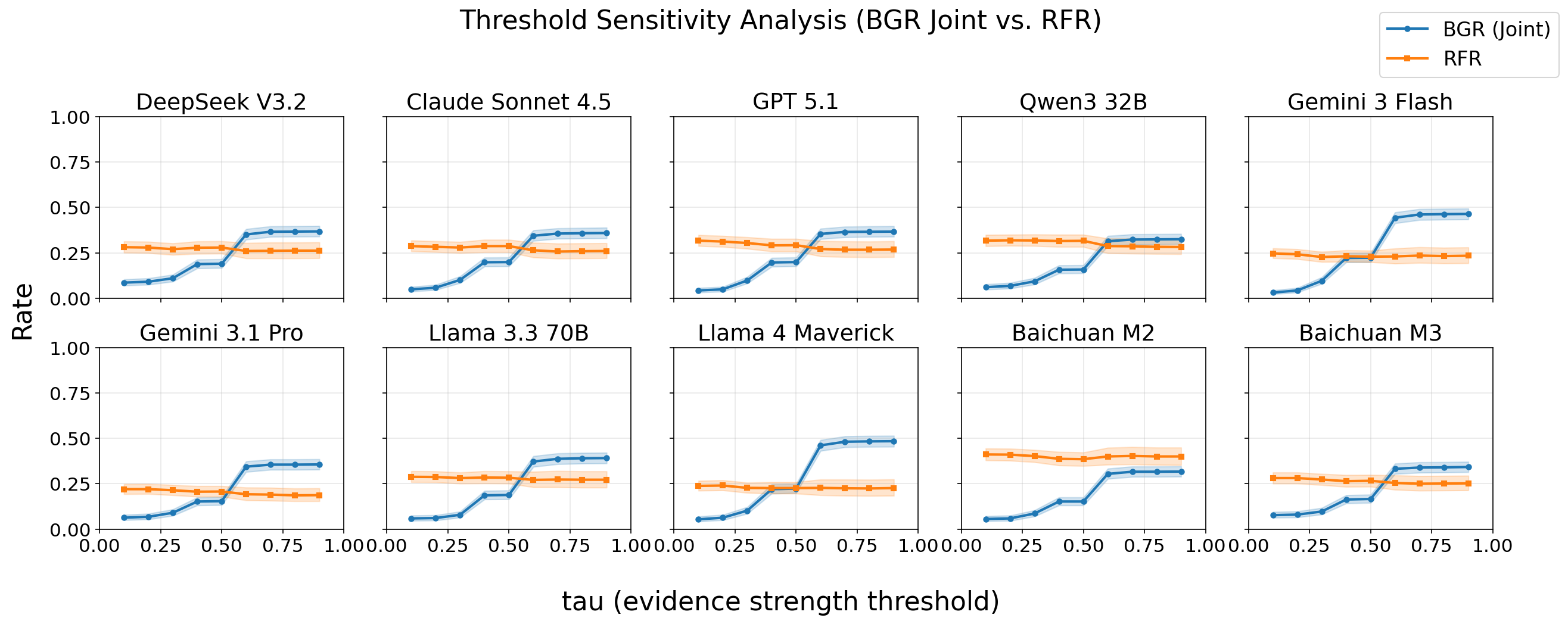}
  \caption{The threshold sensitivity analysis shows BGR and RFR as functions of the evidence strength threshold $\tau$ for all 10 evaluated models.}
  \label{fig:tau_sensitivity}
\end{figure*}

To further investigate the observed evidence-reasoning gap, we conduct a threshold sensitivity analysis by dynamically adjusting the retrieval threshold ($\tau \in [0.1, 0.9]$) and monitoring fluctuations in the Blind Guess Rate (BGR) and Reasoning Failure Rate (RFR).

Figure~\ref{fig:tau_sensitivity} shows two key phenomena. First, BGR increases monotonically with $\tau$ across all models. For instance, Gemini 3 Flash's BGR increases from 3.1\% at $\tau=0.1$ to 46.5\% at $\tau=0.9$, indicating that models rely more on pretrained parametric priors to generate spurious correctness when deprived of explicit contextual grounding. Second, RFR remains stable as the evidence qualification criteria become stricter. Even with near-perfect evidence extraction ($\tau \ge 0.8$), high-performance models (\eg GPT 5.1 and DeepSeek V3.2) still fail to deduce correct outcomes in approximately 26\% of cases. This result indicates that higher-quality context does not resolve the limitations of current LLMs in multi-step symbolic logical synthesis.

\subsection{Extended Analyses}

\subsubsection{End-to-End Complete Consistency}

To assess reliability in practical deployment, we introduce the \textit{Complete Consistency Rate} (Table~\ref{tab:completeness}). This stringent criterion requires zero deviations throughout the hierarchical reasoning chain, from atomic extraction to global aggregation. Note that the near-zero rates for certain models reflect genuine reasoning failures: our pipeline enforces strict JSON formatting and automatically retries malformed outputs, and all reported scores derive from successfully parsed responses (Appendix~\ref{sec:app_format_compliance}).

\begin{table}[t]
\centering
\resizebox{1.0\columnwidth}{!}{%
\addtolength{\tabcolsep}{-0.1em}
\renewcommand{\arraystretch}{1.1}
\begin{tabular}{lcccc}
\toprule
\textbf{Model} & \textbf{Atomic} & \textbf{Domain} & \textbf{Global} & \textbf{E2E} \\
\midrule
\multicolumn{5}{l}{\textit{Proprietary}} \\
GPT 5.1 & 84.03 & 28.12 & 96.17 & 23.80 \\
Claude Sonnet 4.5 & 77.35 & 31.72 & \textbf{100.00} & 29.94 \\
Gemini 3 Flash & \textbf{90.56} & 38.88 & 94.24 & 38.40 \\
Gemini 3.1 Pro & 88.99 & \textbf{46.75} & 99.64 & \textbf{45.13} \\
\midrule
\multicolumn{5}{l}{\textit{Open-weight}} \\
DeepSeek V3.2 & 41.69 & \textbf{37.06} & \textbf{100.00} & \textbf{18.85} \\
Llama 3.3 70B & 0.00 & 5.59 & \textbf{100.00} & 0.00 \\
Qwen3 32B & 8.80 & 3.04 & 99.36 & 1.44 \\
Llama 4 Maverick & 2.10 & 4.03 & 99.84 & 1.29 \\
Baichuan M2 & 23.68 & 2.88 & 99.84 & 0.80 \\
Baichuan M3 & \textbf{46.84} & 15.61 & 99.30 & 11.58 \\
\bottomrule
\end{tabular}%
}
\caption{Complete Consistency Rates (\%).
\textit{Atomic}, \textit{Domain}, and \textit{Global}: The percentage of instances fully consistent with the symbolic rule set within each respective dimension.
\textit{E2E}: The percentage of instances that are fully consistent across the entire end-to-end reasoning chain.}
\label{tab:completeness}
\end{table}

\paragraph{The Error Compounding Effect\textnormal{:}} While models perform relatively well at the atomic level (\eg Gemini 3 Flash at 90.56\%), atomic-level errors compound drastically in multi-step synthesis. Consequently, the state-of-the-art Gemini 3.1 Pro achieves only 45.13\% End-to-End (E2E) Consistency, and high atomic or global answer accuracy creates a dangerous illusion of capability, masking severe intermediate structural fragility.

\subsubsection{Combining Retrieval and Reasoning}

To investigate whether test-time augmentation (TTA) can ameliorate the observed logical misalignment, we conduct ablation studies using CRAG~\citep{yan2024corrective} and Chain-of-Thought (CoT) prompting~\citep{wei2022chain} on three architecturally diverse models: DeepSeek V3.2, Qwen3 32B, and Gemini 3 Flash.

\begin{table}[t]
\centering
\resizebox{\columnwidth}{!}{%
\begin{tabular}{@{}llcccc@{}}
\toprule
& \textbf{Configuration} & \textbf{Jac.} & \textbf{F1} & \textbf{RFR} & \textbf{BGR} \\
\midrule
\multirow{4}{*}{DeepSeek V3.2}
& Baseline            & 54.70 & 61.23 & 26.15 & 36.74 \\
& + CoT               & 54.30 & 60.33 & 24.44 & 33.37 \\
& + CRAG (Agent)      & \textbf{57.47} & \textbf{64.10} & 29.56 & 36.45 \\
& + Agent \& CoT      & 54.20 & 57.94 & \textbf{23.70} & \textbf{26.72} \\
\midrule
\multirow{4}{*}{Qwen3 32B}
& Baseline            & \textbf{61.53} & \textbf{68.16} & 28.32 & 32.44 \\
& + CoT               & 59.84 & 66.58 & 29.00 & 31.11 \\
& + CRAG (Agent)      & 57.73 & 64.45 & 32.75 & 31.36 \\
& + Agent \& CoT      & 58.86 & 62.17 & 30.64 & \textbf{23.85} \\
\midrule
\multirow{4}{*}{Gemini 3 Flash}
& Baseline            & 58.97 & 68.25 & 23.24 & 46.28 \\
& + CoT               & 56.49 & 66.67 & 23.82 & 50.00 \\
& + CRAG (Agent)      & 59.23 & 66.10 & \textbf{19.50} & 40.41 \\
& + Agent \& CoT      & \textbf{62.87} & \textbf{68.73} & 20.24 & \textbf{33.68} \\
\bottomrule
\end{tabular}%
}
\caption{Impact of Tool Augmentation across three models.
\textit{Jac.}/\textit{F1}: Evidence retrieval quality. \textit{RFR}: Reasoning Failure Rate ($\tau{=}0.8$). \textit{BGR}: Blind Guess Rate. Bold indicates the best result for each model and metric.}
\label{tab:ablation}
\end{table}

Results (Table~\ref{tab:ablation}) reveal model-specific TTA response patterns. For DeepSeek V3.2, CRAG expands information boundaries (Jaccard: 57.47) but increases RFR by 3.41 percentage points due to cognitive overload, while CoT reduces BGR by 3.37 percentage points via conservative grounding. Their combination (Agent~\&~CoT) achieves the lowest RFR (23.70\%) and BGR (26.72\%). For Qwen3 32B, CoT alone yields the best RFR (29.00\%), while Agent~\&~CoT drives BGR to its minimum (23.85\%), although retrieval quality degrades under all TTA conditions. For Gemini 3 Flash, CRAG reduces RFR by 3.74 percentage points (to 19.50\%), and the combined Agent~\&~CoT configuration simultaneously maximizes retrieval (Jaccard: 62.87) and minimizes BGR (33.68\%). Across all three models, Agent~\&~CoT consistently produces the lowest BGR, confirming that coupling explicit reasoning with grounded retrieval alleviates spurious correctness regardless of architecture.

\subsubsection{Supplementary Analyses}

Appendix~\ref{sec:app_stats} reports statistical tests across all models, while Appendix~\ref{sec:app_context} analyzes context and retrieval dependencies. We define the \textit{Oracle Evidence Paradox} as the counterintuitive decrease in accuracy when the model is restricted to gold-standard expert sentences rather than the retrieved-context baseline. Gold evidence yields 61.38\% accuracy, 2.84 percentage points below the baseline, possibly because isolated gold sentences omit broader context needed for the final decision. Because this comparison uses a single $T=0$ run on one model and its sample sizes differ by at most two, the result is diagnostic rather than causal. Appendix~\ref{sec:app_neuro_symbolic} shows that rule enforcement improves accuracy by 2--24\% across models. Appendix~\ref{sec:error_taxonomy} shows that domain-level logic misalignment (E2: 29--56\%) is more common than atomic rule violations (E1: 1--32\%), with D2 as the primary bottleneck.

\section{Conclusion}

We introduce \textbf{LogiMed-RoB} and the \textbf{HLC} framework to rigorously audit LLMs against the hierarchical expert logic of the Cochrane RoB~2.0 decision system.
Across 10 models, the \textit{Error Compounding Effect} during multi-step domain synthesis yields near-zero end-to-end consistency for several open-weight architectures.
High Blind Guess Rates and the \textit{Oracle Evidence Paradox}, in which gold evidence underperforms retrieved context, expose evidence-decision decoupling: models can match outcomes without executing clinical deduction.
High final-label accuracy can therefore mask structural fragility, motivating strict, white-box logical verification for clinical deployment.

\section*{Limitations}

While LogiMed-RoB provides a rigorous audit of LLM capabilities in EBM, our study has several limitations. First, our benchmark explicitly focuses on RoB assessment. Although RoB~2.0 represents a quintessential hierarchical expert system, real-world clinical reasoning encompasses a broader spectrum of tasks (\eg diagnostic deduction) that may require adapting our deterministic rule-based mapping to accommodate probabilistic clinical consensus. Second, the current dataset is constructed entirely from English-language RCTs, leaving the cross-lingual reasoning consistency of these architectures unexplored. Third, all models are evaluated under a single-run protocol at temperature $T=0.0$ without variance estimation. While deterministic decoding eliminates sampling stochasticity, it does not account for prompt sensitivity, tokenizer boundary effects, or residual API-side nondeterminism, which is a known issue among several proprietary providers for which identical requests can yield subtly different outputs even at $T=0$. Due to the prohibitive cost of repeated API calls across 10 models and over 15,000 evaluation instances, multi-run statistical robustness analysis was not feasible within our budget constraints. Future work should incorporate repeated trials to quantify result variance. Finally, given the continuous, opaque updates to proprietary architectures, exhaustive prompt optimization and longitudinal tracking remain valuable directions for future work.

\section*{Ethical Considerations}

This research adheres to ethical standards in line with best practices for artificial intelligence and clinical research. While our evaluation framework assists in the automated assessment of bias in randomized controlled trials, it is strictly intended to support rather than replace human expertise and clinical judgment. We advocate for a collaborative model in which technology enhances the efficiency of bias detection, while medical researchers retain full responsibility for interpreting results and making informed decisions.

The complete data collection and review process for the LogiMed-RoB benchmark was conducted exclusively by the authors of this paper. No external personnel or crowd workers were recruited, which ensures strict quality control and eliminates potential labor ethics concerns often associated with extensive dataset annotation. Our approach contributes to improving transparency in clinical trial reporting and encourages the responsible use and further refinement of evaluation tools by the broader research community.

\section*{Acknowledgments}
This work is supported by the National Natural Science Foundation of China (Grant No. 62473271) and the Fundamental Research Funds for the Beijing University of Posts and Telecommunications (Grant No. 2025Al4S03). This work is also supported by the Engineering Research Center of Information Networks, Ministry of Education, China. We would also like to thank the anonymous reviewers and area chairs for constructive discussions and feedback.

\bibliography{references}


\clearpage

\appendix

\section*{Appendix Contents}
\label{sec:appendix_contents}

\noindent To facilitate navigation, the contents of this appendix are organized as follows. Each appendix number links directly to the corresponding section.

\begin{itemize}[leftmargin=*]
    \item \textbf{Appendix~\ref{sec:app_stats}}\textnormal{:} Statistical Significance Framework
    \item \textbf{Appendix~\ref{sec:app_context}}\textnormal{:} Context Configuration Analysis\textnormal{:} Disentangling Retrieval and Reasoning
    \item \textbf{Appendix~\ref{sec:app_neuro_symbolic}}\textnormal{:} Expert Rule Enforcement versus Direct LLM Prediction
    \item \textbf{Appendix~\ref{sec:error_taxonomy}}\textnormal{:} Systematic Error Decomposition
    \item \textbf{Appendix~\ref{sec:app_rob_rules}}\textnormal{:} Cochrane RoB 2.0 Decision Rules
    \item \textbf{Appendix~\ref{sec:app_rob_rules_rob1}}\textnormal{:} Cochrane RoB 1.0 Assessment Structure
    \item \textbf{Appendix~\ref{sec:case_studies}}\textnormal{:} Case Studies
    \item \textbf{Appendix~\ref{sec:app_format_compliance}}\textnormal{:} Output Format Compliance and Validation
    \item \textbf{Appendix~\ref{app:prompt_templates}}\textnormal{:} Prompt Templates
\end{itemize}


\section{Statistical Significance Framework}
\label{sec:app_stats}

To ensure the robustness of the conclusions drawn from the LogiMed-RoB evaluation and to account for stochastic fluctuations in model performance, we establish the following multidimensional hypothesis-testing framework. This framework determines whether the models' performance reflects genuine adherence to medical logic or merely probabilistic shortcut learning and pattern matching.

\subsection{Atomic Consistency\textnormal{:} Binomial Compliance Test}

For Atomic Consistency (CAR), we employ the \textbf{Exact Binomial Test} to determine whether the model's judgments significantly outperform random guessing.

\paragraph{Hypothesis Definition\textnormal{:}} Under the null hypothesis ($H_0$), the probability that the model makes a correct judgment when processing logical branches is $P = \frac{1}{6}$ (equivalent to random guessing over the six-option answer space $\mathcal{A} = \{\text{Y, PY, PN, N, NI, NA}\}$). Under the alternative hypothesis ($H_1$), the probability of a correct judgment is $P \neq \frac{1}{6}$.

\paragraph{Inference Logic\textnormal{:}} If $p < 0.05$ and $\text{CAR} \to 1.0$, the result indicates that the model follows the branch constraints of the expert system. Conversely, $\text{CAR} \to 0$ reflects systematic rule violations by the model.

\subsection{Domain Consistency\textnormal{:} McNemar's Test of Discrepancies}

To examine the phenomenon of ``correct outcome, incorrect logic'' within Domain Consistency, we apply \textbf{McNemar's Test} to the discordant pairs between the model's neural outputs and symbolic derivations.

\paragraph{Hypothesis Definition\textnormal{:}} Let $N_c$ and $N_w$ denote correct and incorrect neural intuition outcomes, respectively. Similarly, let $S_c$ and $S_w$ represent correct and incorrect symbolic derivations. Under $H_0$, we hypothesize that $P(N_c, S_w) = P(N_w, S_c)$. That is, there is no significant difference between the probability of the model exhibiting ``Spurious Understanding'' ($N_c, S_w$) and the probability of ``Over-Adherence to Logic'' ($N_w, S_c$). Under $H_1$, $P(N_c, S_w) \neq P(N_w, S_c)$, indicating significant asymmetry. The direction of this asymmetry indicates which of the two failure modes is more prevalent.

\subsection{Aggregation Consistency\textnormal{:} Systematic Bias Analysis}

For global aggregation logic, accuracy (VR) alone may yield inflated performance estimates under class imbalance. We therefore use \textbf{Quadratic Weighted Cohen's Kappa} for this analysis.

We quantify the level of consistency between the global risk predicted by the neural model, $\hat{R}_{n,\text{global}}^{\text{neural}}$, and the global risk derived from symbolic rules, $R_{n,\text{global}}^{\text{symbolic}}$:

\begin{equation}
\kappa = \frac{p_o - p_e}{1 - p_e},
\end{equation}

\begin{equation}
p_o = \sum_{x \in \mathcal{L}} \sum_{y \in \mathcal{L}} w_{x,y} p_{x,y},
\end{equation}

\begin{equation}
p_e = \sum_{x \in \mathcal{L}} \sum_{y \in \mathcal{L}} w_{x,y} (p_{x \cdot} p_{\cdot y}),
\end{equation}
where the label set $\mathcal{L}$ has an ordinal ranking $\text{Rank}(\cdot) \in \{1, 2, 3\}$ (corresponding to \textit{Low}, \textit{Some Concerns}, and \textit{High}, respectively); $p_{x,y}$ represents the joint probability of a sample receiving a model prediction level of $x$ and a symbolic derivation level of $y$; and $p_{x \cdot}$ and $p_{\cdot y}$ denote their respective marginal probabilities. The weight $w_{x,y} = 1 - \frac{(\text{Rank}(x)-\text{Rank}(y))^2}{(|\mathcal{L}|-1)^2}$ ensures that larger differences in rank receive quadratically greater penalties.

\subsection{Evidential Faithfulness\textnormal{:} Chi-Square Dependency Test}

We use \textbf{Pearson's Chi-square Test} to test the dependence between ``Decision Correctness'' and ``Retrieval Strength (Jaccard)''.

Using a predefined Retrieval Strength threshold $\tau$, the continuous $\text{Jaccard}(A_n, G_n)$ metric is discretized into two levels: ``Strong Evidence Anchoring ($J \ge \tau$)'' and ``Weak Evidence Anchoring ($J < \tau$)''. These levels are cross-tabulated with the binary variable ``Decision Correctness (Correct / Incorrect)'' to construct a $2 \times 2$ contingency table.

\paragraph{Hypothesis Definition\textnormal{:}} Under $H_0$, the correctness of the model's answer is independent of the strength of the supporting clinical evidence that it retrieves. Under $H_1$, the model's decision significantly depends on the accurate retrieval of clinical evidence.

\subsection{Hypothesis Testing Results}
\label{subsec:hypothesis_testing_results}

To verify that the observed consistency patterns reflect systematic logical behaviors rather than stochastic noise, we conducted targeted hypothesis testing across the four evaluation dimensions. The comprehensive statistical results are summarized in Table~\ref{tab:significance}.

\begin{table*}[t]
\centering
\small 
\begin{tabular}{lcccc}
\toprule
\textbf{Model} & \textbf{Atomic} & \textbf{Domain} & \textbf{Agg. $\kappa$ (\%)} & \textbf{Evidential} \\
\midrule
\multicolumn{5}{l}{\textit{Proprietary}} \\
GPT 5.1 & $p < .0001$ & $p < .0001$ & 91.88 & $p < .0001$ \\
Claude Sonnet 4.5 & $p < .0001$ & $p < .0001$ & 100.00 & $p < .0001$ \\
Gemini 3 Flash & $p < .0001$ & $p < .0001$ & 96.45 & $p = .0675$ \\
Gemini 3.1 Pro & $p < .0001$ & $p < .0001$ & 99.73 & $p < .0001$ \\
\midrule
\multicolumn{5}{l}{\textit{Open-weight}} \\
DeepSeek V3.2 & $p < .0001$ & $p < .0001$ & 100.00 & $p < .0001$ \\
Llama 3.3 70B & $p < .0001$ & $p = .0635$ & 100.00 & $p = .0002$ \\
Qwen3 32B & $p < .0001$ & $p < .0001$ & 99.20 & $p < .0001$ \\
Llama 4 Maverick & $p < .0001$ & $p < .0001$ & 99.79 & $p = .0165$ \\
Baichuan M2 & $p < .0001$ & $p < .0001$ & 99.74 & $p = .0059$ \\
Baichuan M3 & $p < .0001$ & $p < .0001$ & 99.44 & $p < .0001$ \\
\bottomrule
\end{tabular}
\caption{Hypothesis Testing Results.
\textit{Atomic}: Exact Binomial Test; \textit{Domain}: McNemar's Test; \textit{Agg.}: Quadratic Weighted Cohen's Kappa multiplied by 100;
\textit{Evidential}: Pearson's $\chi^2$ Test (evaluated at $\tau=0.8$).}
\label{tab:significance}
\end{table*}

\begin{table*}[t]
\centering
\small
\setlength{\tabcolsep}{4pt}
\begin{tabular}{l c c c c c c}
\toprule
\textbf{Model} & $N$ & $b$ ($N_c, S_w$) & $c$ ($N_w, S_c$) & Asymmetry ($b - c$) & $b/c$ & $p$-value \\
\midrule
\multicolumn{7}{l}{\textit{Proprietary}} \\
GPT 5.1 & 3,129 & 974 & 245 & +729 & 3.98 & $< .0001$ \\
Claude Sonnet 4.5 & 3,089 & 721 & 319 & +402 & 2.26 & $< .0001$ \\
Gemini 3 Flash & 3,124 & 735 & 273 & +462 & 2.69 & $< .0001$ \\
Gemini 3.1 Pro & 2,769 & 635 & 170 & +465 & 3.74 & $< .0001$ \\
\midrule
\multicolumn{7}{l}{\textit{Open-weight}} \\
DeepSeek V3.2 & 3,129 & 1,110 & 240 & +870 & 4.63 & $< .0001$ \\
Llama 3.3 70B & 3,125 & 855 & 779 & +76 & 1.10 & .0635 \\
Qwen3 32B & 3,119 & 759 & 976 & $-$217 & 0.78 & $< .0001$ \\
Llama 4 Maverick & 3,097 & 642 & 797 & $-$155 & 0.81 & $< .0001$ \\
Baichuan M2 & 3,124 & 911 & 645 & +266 & 1.41 & $< .0001$ \\
Baichuan M3 & 2,849 & 683 & 439 & +244 & 1.56 & $< .0001$ \\
\bottomrule
\end{tabular}
\caption{McNemar's Test asymmetry analysis. $b$ ($N_c, S_w$): correct neural output but incorrect symbolic derivation (Spurious Understanding). $c$ ($N_w, S_c$): incorrect neural output but correct symbolic derivation (Over-Adherence to Logic). Positive asymmetry indicates that Spurious Understanding dominates. Negative asymmetry indicates that Over-Adherence to Logic dominates.}
\label{tab:mcnemar_asymmetry}
\end{table*}

\paragraph{Atomic Consistency\textnormal{:}} Exact binomial tests strongly reject the null hypothesis of random guessing ($H_0: P=\frac{1}{6}, p < 0.0001$) across all models. This confirms genuine rule adherence at the micro-level, indicating that models reliably process foundational signaling questions rather than relying on stochastic outputs.

\paragraph{Domain Consistency\textnormal{:}} McNemar's test reveals pervasive rule-output misalignment at the domain level. As shown in Table~\ref{tab:mcnemar_asymmetry}, the direction and magnitude of asymmetry vary across models. For seven models (GPT 5.1, Claude Sonnet 4.5, Gemini 3 Flash, Gemini 3.1 Pro, DeepSeek V3.2, and the two Baichuan models), $b \gg c$ ($b/c$ ratios of 1.41--4.63), indicating that \textit{Spurious Understanding}, in which models arrive at correct outcomes despite flawed symbolic reasoning, is the dominant failure mode. Conversely, Qwen3 32B and Llama 4 Maverick exhibit reverse asymmetry ($c > b$, $b/c$ = 0.78--0.81), in which models produce incorrect outcomes despite correct symbolic derivations (\textit{Over-Adherence to Logic}). The sole non-significant result, Llama 3.3 70B ($p = .064$, $b/c = 1.10$, asymmetry = +76), reflects near-symmetric discordant pairs ($b \approx c$) that statistically cancel out, rather than indicating robust logical binding.

\paragraph{Aggregation Consistency\textnormal{:}} For global aggregation, near-perfect Quadratic Weighted Cohen's Kappa scores (ranging from 91.88 to 100.00) demonstrate that LLMs lack systematic directional bias during macro-level extremum synthesis.

\paragraph{Evidential Faithfulness\textnormal{:}} At $\tau=0.8$, Pearson's $\chi^2$ tests reject independence between evidence retrieval and decision accuracy for nine of the 10 models at $\alpha=0.05$. Gemini 3 Flash is the exception ($p=.0675$). Statistical dependence does not establish evidential causality. The high Blind Guess Rates reported in the main text show that correct decisions can still occur without strong evidence overlap, revealing an evidence-reasoning gap in current architectures.


\section{Context Configuration Analysis\textnormal{:} Disentangling Retrieval and Reasoning}
\label{sec:app_context}

To isolate the root cause of the logical consistency gap, we decouple evidence extraction from logical decision-making by evaluating three configurations: (1) \textit{Window + Retrieval (Baseline)}, (2) \textit{Full Window, No Retrieval}, and (3) \textit{Gold Evidence Only}.

\begin{table*}[t]
\centering
\small 
\setlength{\tabcolsep}{10pt}
\begin{tabular}{l c cccc}
\toprule
\textbf{Configuration} & $N$ & \textbf{Overall Accuracy} & \textbf{Low} & \textbf{Some Concerns} & \textbf{High} \\
\midrule
Window + Retrieval (Baseline) & 1,048 & \textbf{64.22} & \textbf{68.31} & 47.37 & 60.69 \\
Full Window, No Retrieval & 1,047 & 63.71 & 67.07 & 49.34 & \textbf{61.38} \\
Gold Evidence Only & 1,046 & 61.38 & 63.07 & \textbf{54.30} & 60.00 \\
\bottomrule
\end{tabular}
\caption{Reasoning performance under varying evidence context configurations.}
\label{tab:context}
\end{table*}

The comparative analysis (Table~\ref{tab:context}) yields three critical findings:

\begin{itemize}[leftmargin=*]
    \item \textbf{Retrieval Provides Modest Guidance}\textnormal{:} The baseline slightly outperforms the full-window, non-retrieval setup (64.22\% vs.\ 63.71\%), a difference of 0.51 percentage points. This result is consistent with explicit citation instructions providing a modest inductive bias that grounds the model's attention.
    
    \item \textbf{Dichotomy of Positive vs.\ Missing Logic}\textnormal{:} Removing the retrieval requirement reduces ``Low'' risk accuracy by 1.24 percentage points while increasing ``Some Concerns'' accuracy by 1.97 percentage points and ``High'' risk accuracy by 0.69 percentage points. These small, class-dependent changes suggest that retrieval instructions affect categories differently.
    
    \item \textbf{The Oracle Evidence Paradox}\textnormal{:} Restricting the model to gold-standard expert evidence yields 61.38\% accuracy, which is 2.84 percentage points below the baseline and 2.33 percentage points below the full-window, non-retrieval configuration. One plausible explanation is that the gold sentences omit broader contextual information needed for the final decision. Because these results come from a single $T=0$ run on one model and the number of valid instances differs by at most two, they should be interpreted as diagnostic rather than causal evidence. Together with the Reasoning Failure Rates of 18.63--40.05\%, the result indicates that higher-quality evidence alone does not guarantee correct multi-step synthesis.
\end{itemize}


\section{Expert Rule Enforcement versus Direct LLM Prediction}
\label{sec:app_neuro_symbolic}

To directly demonstrate the value of enforcing the Cochrane expert rules as a post-hoc verification layer, we conduct a comparative experiment contrasting two paradigms: (1)~\textbf{Rule-Guided Deduction (Logical Accuracy)}, in which the model's atomic-level outputs are fed into the deterministic Cochrane decision map $\mathcal{M}_i$ to derive domain-level risk labels, and (2)~\textbf{Direct LLM Prediction (LLM Accuracy)}, in which the model directly generates domain-level risk labels without any rule-based post-processing.

\subsection{Experimental Setup}

For each model and each RoB~2.0 domain, we compute two accuracy scores over the same evaluation set. \textit{Logical Accuracy} is obtained by (a)~prompting the model to answer all signaling questions for a given domain, (b)~applying the Cochrane decision map to the model's atomic answers to deterministically derive the domain risk label, and (c)~comparing this derived label against the expert gold standard. \textit{LLM Accuracy} is obtained by directly prompting the model for a domain-level risk judgment and comparing its output against the gold standard. Both metrics are reported with 95\% confidence intervals computed using the Wilson score method.

\subsection{Results and Analysis}

\begin{table*}[!t]
\centering
\small
\setlength{\tabcolsep}{3.0pt}
\begin{tabular}{l l cc cc}
\toprule
\textbf{Model} & \textbf{Domain} & \textbf{Logical Acc.} & \textbf{95\% CI} & \textbf{LLM Acc.} & \textbf{95\% CI} \\
\midrule
\multirow{6}{*}{Qwen3 32B}
& D1 & 0.5753 & [0.5362, 0.6135] & 0.5897 & [0.5507, 0.6277] \\
& D2 & 0.5392 & [0.5000, 0.5779] & 0.3024 & [0.2677, 0.3395] \\
& D3 & 0.4800 & [0.4411, 0.5192] & 0.3232 & [0.2877, 0.3609] \\
& D4 & 0.7984 & [0.7650, 0.8281] & 0.2839 & [0.2498, 0.3206] \\
& D5 & 0.5264 & [0.4872, 0.5653] & 0.2432 & [0.2112, 0.2783] \\
& \textbf{Overall} & \textbf{0.5835} & [0.5661, 0.6007] & 0.3485 & [0.3320, 0.3654] \\
\midrule
\multirow{6}{*}{Llama 3.3 70B}
& D1 & 0.6827 & [0.6452, 0.7180] & 0.6923 & [0.6550, 0.7273] \\
& D2 & 0.5904 & [0.5514, 0.6283] & 0.1520 & [0.1260, 0.1823] \\
& D3 & 0.3968 & [0.3592, 0.4357] & 0.1408 & [0.1157, 0.1703] \\
& D4 & 0.8320 & [0.8007, 0.8593] & 0.5616 & [0.5224, 0.6000] \\
& D5 & 0.3818 & [0.3446, 0.4205] & 0.2923 & [0.2581, 0.3291] \\
& \textbf{Overall} & \textbf{0.5766} & [0.5592, 0.5939] & 0.3677 & [0.3509, 0.3847] \\
\midrule
\multirow{6}{*}{DeepSeek V3.2}
& D1 & 0.5435 & [0.4899, 0.5962] & 0.5826 & [0.5290, 0.6343] \\
& D2 & 0.4925 & [0.4392, 0.5460] & 0.4505 & [0.3979, 0.5042] \\
& D3 & 0.4895 & [0.4362, 0.5430] & 0.5375 & [0.4839, 0.5904] \\
& D4 & 0.5886 & [0.5350, 0.6401] & 0.5015 & [0.4481, 0.5549] \\
& D5 & 0.4595 & [0.4067, 0.5131] & 0.4835 & [0.4303, 0.5370] \\
& \textbf{Overall} & \textbf{0.5147} & [0.4907, 0.5387] & 0.5111 & [0.4871, 0.5351] \\
\midrule
\multirow{6}{*}{Claude Sonnet 4.5}
& D1 & 0.5906 & [0.5514, 0.6287] & 0.5987 & [0.5596, 0.6366] \\
& D2 & 0.6634 & [0.6253, 0.6996] & 0.5243 & [0.4849, 0.5634] \\
& D3 & 0.6343 & [0.5956, 0.6713] & 0.6570 & [0.6187, 0.6933] \\
& D4 & 0.7650 & [0.7300, 0.7967] & 0.7472 & [0.7114, 0.7799] \\
& D5 & 0.6570 & [0.6187, 0.6933] & 0.5340 & [0.4946, 0.5730] \\
& \textbf{Overall} & \textbf{0.6620} & [0.6452, 0.6785] & 0.6122 & [0.5949, 0.6292] \\
\bottomrule
\end{tabular}
\caption{Comparison of Rule-Guided Deduction (\textit{Logical Accuracy}) versus Direct LLM Prediction (\textit{LLM Accuracy}) with 95\% confidence intervals. \textit{Logical Accuracy} is derived by feeding each model's atomic outputs into the Cochrane decision rules. \textit{LLM Accuracy} reflects the model's direct, unguided domain-level predictions.}
\label{tab:neuro_symbolic}
\end{table*}

\begin{table*}[!t]
\ContinuedFloat
\centering
\small
\setlength{\tabcolsep}{3.0pt}
\begin{tabular}{l l cc cc}
\toprule
\textbf{Model} & \textbf{Domain} & \textbf{Logical Acc.} & \textbf{95\% CI} & \textbf{LLM Acc.} & \textbf{95\% CI} \\
\midrule
\multirow{6}{*}{GPT 5.1}
& D1 & 0.6502 & [0.6120, 0.6865] & 0.5431 & [0.5040, 0.5818] \\
& D2 & 0.4569 & [0.4182, 0.4960] & 0.6502 & [0.6120, 0.6865] \\
& D3 & 0.5256 & [0.4864, 0.5644] & 0.4920 & [0.4530, 0.5311] \\
& D4 & 0.8064 & [0.7736, 0.8355] & 0.8048 & [0.7719, 0.8340] \\
& D5 & 0.3339 & [0.2980, 0.3717] & 0.2780 & [0.2443, 0.3143] \\
& \textbf{Overall} & \textbf{0.5545} & [0.5370, 0.5718] & 0.5535 & [0.5361, 0.5709] \\
\midrule
\multirow{6}{*}{Gemini 3 Flash}
& D1 & 0.6544 & [0.6163, 0.6906] & 0.6768 & [0.6391, 0.7123] \\
& D2 & 0.6944 & [0.6572, 0.7292] & 0.5232 & [0.4840, 0.5621] \\
& D3 & 0.6384 & [0.6000, 0.6751] & 0.5952 & [0.5562, 0.6330] \\
& D4 & 0.7949 & [0.7614, 0.8247] & 0.7853 & [0.7513, 0.8157] \\
& D5 & 0.6736 & [0.6359, 0.7092] & 0.6304 & [0.5919, 0.6673] \\
& \textbf{Overall} & \textbf{0.6911} & [0.6747, 0.7071] & 0.6421 & [0.6252, 0.6588] \\
\midrule
\multirow{6}{*}{Gemini 3.1 Pro}
& D1 & 0.7004 & [0.6609, 0.7370] & 0.7112 & [0.6721, 0.7474] \\
& D2 & 0.6751 & [0.6350, 0.7128] & 0.5614 & [0.5198, 0.6021] \\
& D3 & 0.6336 & [0.5927, 0.6726] & 0.6029 & [0.5616, 0.6428] \\
& D4 & 0.8409 & [0.8080, 0.8690] & 0.8354 & [0.8022, 0.8640] \\
& D5 & 0.6913 & [0.6517, 0.7284] & 0.6859 & [0.6461, 0.7232] \\
& \textbf{Overall} & \textbf{0.7082} & [0.6910, 0.7248] & 0.6793 & [0.6617, 0.6964] \\
\midrule
\multirow{6}{*}{Llama 4 Maverick}
& D1 & 0.6446 & [0.6061, 0.6813] & 0.6559 & [0.6176, 0.6923] \\
& D2 & 0.5654 & [0.5261, 0.6040] & 0.2181 & [0.1874, 0.2523] \\
& D3 & 0.6226 & [0.5838, 0.6599] & 0.1629 & [0.1359, 0.1940] \\
& D4 & 0.7932 & [0.7595, 0.8233] & 0.6656 & [0.6275, 0.7016] \\
& D5 & 0.7016 & [0.6644, 0.7363] & 0.6194 & [0.5805, 0.6567] \\
& \textbf{Overall} & \textbf{0.6655} & [0.6487, 0.6819] & 0.4643 & [0.4468, 0.4819] \\
\bottomrule
\end{tabular}
\caption[]{Comparison of Rule-Guided Deduction versus Direct LLM Prediction (continued).}
\end{table*}

Table~\ref{tab:neuro_symbolic} presents the complete results across all eight models and five RoB~2.0 domains. Expert rule enforcement consistently improves over direct prediction: the overall Logical Accuracy exceeds LLM Accuracy in seven of eight models (the sole exception being DeepSeek V3.2, where the two are virtually tied at 51.47\% vs.\ 51.11\%). The improvement is most dramatic for open-weight models. Llama 4 Maverick gains 20.12 percentage points, Qwen3 32B gains 23.50 percentage points, and Llama 3.3 70B gains 20.89 percentage points, while even the strongest proprietary model, Gemini 3.1 Pro, benefits by 2.89 percentage points. The 95\% CIs for Logical Accuracy and LLM Accuracy do not overlap for most model--domain pairs.

The benefit is domain-dependent. Domain~D1 (Randomization Process) shows minimal or even negative gains for most models. For instance, GPT 5.1 drops from 0.6502 to 0.5431 on D1, likely because D1's signaling questions are relatively factual and straightforward, allowing the LLM to make direct judgments effectively. In contrast, Domains~D2 and D4 exhibit the largest improvements: for Llama 3.3 70B on D4, Logical Accuracy reaches 83.20\% compared to a mere 56.16\% LLM Accuracy, a gap of +27.04\%. This pattern aligns with our earlier finding that D2 and D4 contain longer, more complex decision chains where rule-based deduction is essential.


\section{Systematic Error Decomposition}
\label{sec:error_taxonomy}

To address the need for fine-grained failure diagnosis, we decompose model errors into a four-level taxonomy aligned with the HLC framework. Each level targets a distinct reasoning failure mode, enabling precise localization of \textit{where} and \textit{how} models deviate from expert logic.

\subsection{Error Classification Framework}

\paragraph{E1\textnormal{:} Atomic Rule Violations}
These errors occur at the signaling question level when the model's answer violates the branch constraints of the RoB~2.0 decision tree (\eg by answering a downstream question with a value inconsistent with a prerequisite answer). They are measured via $1 - \text{CAR}$ across domains D2--D4.

\paragraph{E2\textnormal{:} Domain Logic Misalignment}
These errors are McNemar discordant pairs in which the model's neural output and the symbolic derivation disagree. We distinguish two asymmetric failure modes:
\begin{itemize}[leftmargin=*]
    \item \textbf{Nc,Sw (Spurious Understanding)}\textnormal{:} The model predicts the correct domain label, but the rule-based derivation from its own atomic answers yields a different label. The model ``gets the right answer for the wrong reasons''.
    \item \textbf{Nw,Sc (Over-Adherence to Logic)}\textnormal{:} The model predicts incorrectly, yet its atomic answers, when processed through the decision map, produce the correct label. The model's surface-level judgment overrides its own logical evidence.
\end{itemize}

\paragraph{E3\textnormal{:} Aggregation Failures}
These errors violate the Worst-of Principle: domain-level labels are correctly derived, but the global risk judgment is incorrect. They are rare (most models achieve $>$99\% VR), indicating that LLMs readily internalize simple extremum rules.

\paragraph{E4\textnormal{:} Evidential Failures}
We define two complementary failure modes under $\tau{=}0.8$:
\begin{itemize}[leftmargin=*]
    \item \textbf{E4a (Blind Guess, BGR)}\textnormal{:} Jaccard similarity $< \tau$, yet the prediction is correct, indicating spurious correctness via parametric priors rather than evidence grounding.
    \item \textbf{E4b (Reasoning Failure, RFR)}\textnormal{:} Jaccard similarity $\ge \tau$, yet the prediction is wrong, indicating that the model retrieves sufficient evidence but fails to reason over it.
\end{itemize}

\subsection{Error Distribution Across Models}

Table~\ref{tab:error_taxonomy} presents the error counts and percentages across all 10 models. Three key patterns emerge:

\paragraph{Domain logic misalignment (E2) is the dominant error source.}
E2 accounts for 29--56\% of all cases, dwarfing atomic errors (E1: 1--32\%) and aggregation errors (E3: $<$6\%). This confirms that the primary bottleneck lies not in individual signaling question processing but in the multi-step deduction from atomic answers to domain-level risk labels.

\paragraph{Error direction reveals model capability tiers.}
The Nc,Sw-to-Nw,Sc ratio stratifies models into three tiers. \textit{Tier~1} (proprietary models and DeepSeek): Nc,Sw dominates with a ratio of 2.3--4.6$\times$, indicating that these models frequently arrive at correct domain labels through flawed internal logic. In other words, they obtain the ``right answer for the wrong reasons''. \textit{Tier~2} (Baichuan models): Nc,Sw shows moderate dominance (1.4--1.6$\times$). \textit{Tier~3} (Qwen3 32B and Llama 4 Maverick): Nw,Sc dominates (ratio $<$1.0), indicating that these models' direct judgments are worse than the judgments produced by their own atomic logic.

\paragraph{E4b (Reasoning Failure) is uniformly high.}
RFR ranges from 18.6\% (Gemini 3.1 Pro) to 40.0\% (Baichuan M2), confirming that even with high-quality evidence retrieval, current LLMs cannot reliably translate extracted facts into correct logical derivations.

\subsection{Reasoning Stage Failure Localization}

Table~\ref{tab:stage_failure} identifies each model's bottleneck stage, which is the reasoning level with the highest failure rate. A clear bifurcation emerges:
\begin{itemize}[leftmargin=*]
    \item \textbf{Proprietary models} exhibit a bottleneck at the evidential stage (RFR 18.6--26.7\%): they can process signaling questions and derive domain labels reasonably well but struggle to ground decisions in retrieved evidence.
    \item \textbf{Open-weight models} exhibit a bottleneck at the domain stage (failure rate 27.4--48.6\%): they fail earlier and are unable to aggregate atomic signals into coherent domain-level conclusions.
\end{itemize}

The error compounding rate (Domain Error\% / Atomic Error\%) quantifies this amplification. For proprietary models, domain errors occur at 7--14$\times$ the rate of atomic errors. For weaker open-weight models, the ratio drops to 1.3--2.5$\times$. This lower ratio does not indicate less compounding. Instead, these models' atomic error rates are already high (19--32\%), leaving less room for amplification.

\subsection{Domain-Specific Failure Patterns}

Table~\ref{tab:mcnemar_detail} presents the McNemar discordant pairs decomposed by RoB~2.0 domain, and Table~\ref{tab:domain_error} cross-references these with Logical Fidelity.

\paragraph{D2 is universally the hardest domain and exhibits a unique error direction.}
D2 has the lowest LF for eight of the 10 models and remains low across all models (22.7--71.7\%). Critically, D2 is the \textit{only} domain where Nw,Sc $>$ Nc,Sw for nine of the 10 models, meaning that the models' rule-derived labels tend to be correct while their direct domain predictions are wrong. This ``Over-Adherence to Logic'' pattern is unique to D2 and stems from its complex decision map involving per-protocol versus intention-to-treat comparisons, where models' surface-level pattern matching diverges from the required multi-step conditional logic.

\paragraph{CAR-to-LF gap varies dramatically by domain.}
For top proprietary models, the gap between atomic accuracy (CAR) and domain fidelity (LF) is 41--50\% for D2, 7--16\% for D3, and 2--10\% for D4. This gradient reveals that D2's decision map introduces the greatest logical complexity, while D4's relatively straightforward deduction chain preserves atomic-level accuracy. Notably, DeepSeek V3.2 shows a more uniform gap across domains (D2: 24\%, D3: 22\%, and D4: 8\%), suggesting that its errors are more evenly distributed rather than concentrated in specific logical transitions.

\paragraph{D5 (Selection of Reported Result) shows capability-dependent patterns.}
Strong models show large positive $\Delta$ (GPT 5.1: +373), while weaker models show negative $\Delta$ (Qwen3 32B: $-$26, Baichuan M2: $-$38). This indicates that D5's evaluation criteria, which require assessing selective reporting across multiple outcomes, challenge weaker models' ability to maintain logical consistency, while strong models can leverage their broader reasoning capacity.

\subsection{Evidential Grounding Analysis}

Table~\ref{tab:evidence_quality} presents the average Jaccard similarity for correctly and incorrectly predicted samples. Across all 10 models, correctly predicted samples consistently exhibit higher retrieval quality (gap: 0.09--0.24), confirming a statistical dependence between evidence grounding and decision accuracy. However, this dependence is weaker than expected: even incorrect predictions show substantial retrieval (J\,=\,0.42--0.52), indicating that models can access relevant evidence but fail to reason over it correctly. The conditional Blind Guess Rate (51--72\% of low-evidence predictions are correct) further confirms heavy reliance on parametric priors rather than evidence-based deduction.

\begin{table*}[t]
\centering
\small
\setlength{\tabcolsep}{4pt}
\begin{tabular}{l r@{\,}r r@{\,}r r@{\,}r r@{\,}r}
\toprule
\textbf{Model} & \multicolumn{2}{c}{\textbf{E1: Atomic}} & \multicolumn{2}{c}{\textbf{E2: Domain}} & \multicolumn{2}{c}{\textbf{E3: Agg.}} & \multicolumn{2}{c}{\textbf{E4: Evidential}} \\
\cmidrule(lr){2-3} \cmidrule(lr){4-5} \cmidrule(lr){6-7} \cmidrule(lr){8-9}
 & \multicolumn{2}{c}{(rule violations)} & \multicolumn{2}{c}{(logic misalignment)} & \multicolumn{2}{c}{(worst-of errors)} & \multicolumn{2}{c}{(BGR\,/\,RFR)} \\
\midrule
GPT 5.1             & 104  & (\textit{1.7}\%) & 1,219 & (39.0\%) & 24 & (3.8\%) & 36.6\% & /\,26.7\% \\
Claude Sonnet 4.5   & 173  & (2.8\%) & 1,040 & (33.7\%) & 0  & (0.0\%) & 35.8\% & /\,25.9\% \\
Gemini 3 Flash       & 72   & (\textit{1.2}\%) & 1,008 & (32.3\%) & 36 & (5.8\%) & 46.3\% & /\,23.2\% \\
Gemini 3.1 Pro       & 62   & (\textbf{1.1}\%) & 805  & (\textbf{29.1}\%) & 2  & (\textit{0.4}\%) & 35.5\% & /\,\textbf{18.6}\% \\
\midrule
DeepSeek V3.2       & 500  & (8.0\%) & 1,350 & (43.1\%) & 0  & (0.0\%) & 36.7\% & /\,26.2\% \\
Llama 3.3 70B       & 1,992 & (31.8\%) & 1,634 & (52.3\%) & 0  & (0.0\%) & 39.0\% & /\,27.2\% \\
Qwen3 32B           & 1,198 & (19.2\%) & 1,735 & (55.6\%) & 4  & (0.6\%) & \textbf{32.4}\% & /\,28.3\% \\
Llama 4 Maverick    & 1,674 & (27.0\%) & 1,439 & (46.5\%) & 1  & (0.2\%) & 48.3\% & /\,22.4\% \\
Baichuan M2         & 884  & (14.1\%) & 1,556 & (49.8\%) & 1  & (0.2\%) & \textit{31.6}\% & /\,40.0\% \\
Baichuan M3         & 341  & (6.0\%)  & 1,122 & (39.4\%) & 4  & (0.7\%) & 34.0\% & /\,25.1\% \\
\bottomrule
\end{tabular}
\caption{Systematic error taxonomy across the HLC framework.
\textbf{E1}: signaling question rule violations (D2--D4).
\textbf{E2}: McNemar discordant pairs across all five domains (Nc,Sw\,+\,Nw,Sc).
\textbf{E3}: Worst-of Principle aggregation failures.
\textbf{E4}: Blind Guess Rate (BGR) and Reasoning Failure Rate (RFR) at $\tau{=}0.8$.
Bold indicates the best result in each column. Italics indicate the second-best result.}
\label{tab:error_taxonomy}
\end{table*}

\begin{table}[t]
\centering
\small
\setlength{\tabcolsep}{4pt}
\begin{tabular}{l cccc}
\toprule
\textbf{Model} & \textbf{Atomic} & \textbf{Domain} & \textbf{Agg.} & \textbf{Evid.} \\
\midrule
GPT 5.1             & 1.7\%  & 21.3\% & 3.8\%  & \textbf{26.7\%} \\
Claude Sonnet 4.5   & 2.8\%  & 20.8\% & 0.0\%  & \textbf{25.9\%} \\
Gemini 3 Flash       & 1.2\%  & 16.1\% & 5.8\%  & \textbf{23.2\%} \\
Gemini 3.1 Pro       & 1.1\%  & 12.4\% & 0.4\%  & \textbf{18.6\%} \\
\midrule
DeepSeek V3.2       & 8.0\%  & 20.0\% & 0.0\%  & \textbf{26.2\%} \\
Llama 3.3 70B       & 31.8\% & \textbf{39.9\%} & 0.0\% & 27.2\% \\
Qwen3 32B           & 19.2\% & \textbf{48.6\%} & 0.6\% & 28.3\% \\
Llama 4 Maverick    & 27.0\% & \textbf{38.5\%} & 0.2\% & 22.4\% \\
Baichuan M2         & 14.1\% & \textbf{43.1\%} & 0.2\% & 40.0\% \\
Baichuan M3         & 6.0\%  & \textbf{27.4\%} & 0.7\% & 25.1\% \\
\bottomrule
\end{tabular}
\caption{Reasoning stage failure rates (\%).
\textit{Atomic}: $1 - \text{CAR}$.
\textit{Domain}: $1 - \text{LF}$.
\textit{Agg.}: $1 - \text{VR}$.
\textit{Evid.}: RFR at $\tau{=}0.8$.
Bold marks each model's bottleneck stage.}
\label{tab:stage_failure}
\end{table}

\begin{table*}[t]
\centering
\footnotesize
\setlength{\tabcolsep}{2.5pt}
\resizebox{\textwidth}{!}{%
\begin{tabular}{l ccc ccc ccc ccc ccc}
\toprule
 & \multicolumn{3}{c}{\textbf{D1}} & \multicolumn{3}{c}{\textbf{D2}} & \multicolumn{3}{c}{\textbf{D3}} & \multicolumn{3}{c}{\textbf{D4}} & \multicolumn{3}{c}{\textbf{D5}} \\
\cmidrule(lr){2-4} \cmidrule(lr){5-7} \cmidrule(lr){8-10} \cmidrule(lr){11-13} \cmidrule(lr){14-16}
\textbf{Model} & Nc,Sw & Nw,Sc & $\Delta$ & Nc,Sw & Nw,Sc & $\Delta$ & Nc,Sw & Nw,Sc & $\Delta$ & Nc,Sw & Nw,Sc & $\Delta$ & Nc,Sw & Nw,Sc & $\Delta$ \\
\midrule
GPT 5.1           & 183 & 18  & +165 & 91  & 196 & $-$105 & 215 & 15  & +200 & 101 & 5   & +96  & 384 & 11  & +373 \\
Claude Sonnet 4.5 & 193 & 10  & +183 & 95  & 163 & $-$68  & 160 & 8   & +152 & 97  & 39  & +58  & 176 & 99  & +77  \\
Gemini 3 Flash     & 182 & 10  & +172 & 69  & 189 & $-$120 & 199 & 32  & +167 & 117 & 8   & +109 & 168 & 34  & +134 \\
Gemini 3.1 Pro     & 142 & 3   & +139 & 79  & 129 & $-$50  & 163 & 27  & +136 & 81  & 7   & +74  & 170 & 4   & +166 \\
\midrule
DeepSeek V3.2     & 211 & 13  & +198 & 163 & 102 & +61  & 161 & 33  & +128 & 271 & 75  & +196 & 304 & 17  & +287 \\
Llama 3.3 70B     & 148 & 7   & +141 & 91  & 318 & $-$227 & 250 & 212 & +38  & 57  & 178 & $-$121 & 309 & 64  & +245 \\
Qwen3 32B         & 209 & 18  & +191 & 162 & 191 & $-$29  & 127 & 171 & $-$44 & 43  & 352 & $-$309 & 218 & 244 & $-$26 \\
Llama 4 Maverick  & 186 & 15  & +171 & 188 & 245 & $-$57  & 83  & 351 & $-$268 & 61 & 104 & $-$43 & 124 & 82  & +42  \\
Baichuan M2       & 210 & 23  & +187 & 187 & 235 & $-$48  & 206 & 81  & +125 & 105 & 65  & +40  & 203 & 241 & $-$38 \\
Baichuan M3       & 184 & 9   & +175 & 99  & 251 & $-$152 & 180 & 18  & +162 & 66  & 13  & +53  & 154 & 148 & +6   \\
\bottomrule
\end{tabular}%
}
\caption{McNemar discordant pairs per RoB~2.0 domain.
Nc,Sw: model correct but rule derivation wrong (\textit{Spurious Understanding}).
Nw,Sc: model wrong but rule derivation correct (\textit{Over-Adherence to Logic}).
$\Delta = \text{Nc,Sw} - \text{Nw,Sc}$: positive values indicate that Spurious Understanding predominates.
D2 exhibits negative $\Delta$ for nine of the 10 models, identifying it as the primary domain in which models produce incorrect direct judgments despite correct rule derivations.}
\label{tab:mcnemar_detail}
\end{table*}

\begin{table*}[t]
\centering
\small
\setlength{\tabcolsep}{4pt}
\begin{tabular}{l cc cc cc cc cc}
\toprule
 & \multicolumn{2}{c}{\textbf{D1}} & \multicolumn{2}{c}{\textbf{D2}} & \multicolumn{2}{c}{\textbf{D3}} & \multicolumn{2}{c}{\textbf{D4}} & \multicolumn{2}{c}{\textbf{D5}} \\
\cmidrule(lr){2-3} \cmidrule(lr){4-5} \cmidrule(lr){6-7} \cmidrule(lr){8-9} \cmidrule(lr){10-11}
\textbf{Model} & LF & Err & LF & Err & LF & Err & LF & Err & LF & Err \\
\midrule
GPT 5.1             & 80.7\% & 201 & 48.2\% & 287 & 81.2\% & 230 & 95.8\% & 106 & 87.4\% & 395 \\
Claude Sonnet 4.5   & 88.7\% & 203 & 55.3\% & 258 & 88.0\% & 168 & 85.9\% & 136 & 78.2\% & 275 \\
Gemini 3 Flash       & 93.0\% & 192 & 50.2\% & 258 & 90.6\% & 231 & 97.0\% & 125 & 88.8\% & 202 \\
Gemini 3.1 Pro       & 95.1\% & 145 & 58.5\% & 208 & 87.9\% & 190 & 97.5\% & 88  & 99.1\% & 174 \\
\midrule
DeepSeek V3.2       & 91.9\% & 224 & 71.7\% & 265 & 72.5\% & 194 & 76.5\% & 346 & 87.4\% & 321 \\
Llama 3.3 70B       & 90.9\% & 155 & 22.7\% & 409 & 45.8\% & 462 & 63.8\% & 235 & 77.3\% & 373 \\
Qwen3 32B           & 88.1\% & 227 & 49.3\% & 353 & 41.0\% & 298 & 30.0\% & 395 & 48.5\% & 462 \\
Llama 4 Maverick    & 92.1\% & 201 & 47.3\% & 433 & 19.0\% & 434 & 72.4\% & 165 & 76.9\% & 206 \\
Baichuan M2         & 82.9\% & 233 & 41.6\% & 422 & 49.9\% & 287 & 64.7\% & 170 & 45.4\% & 444 \\
Baichuan M3         & 89.8\% & 193 & 35.3\% & 350 & 84.4\% & 198 & 89.5\% & 79  & 63.9\% & 302 \\
\bottomrule
\end{tabular}
\caption{Domain-level Logical Fidelity (LF) and total McNemar errors (Err\,=\,Nc,Sw\,+\,Nw,Sc) per domain.
D2 exhibits the lowest LF for eight of the 10 models, confirming it as the most frequent logical bottleneck.}
\label{tab:domain_error}
\end{table*}

\begin{table}[t]
\centering
\small
\begin{tabular}{l ccc}
\toprule
\textbf{Model} & $\bar{J}_{\text{correct}}$ & $\bar{J}_{\text{wrong}}$ & $\Delta J$ \\
\midrule
GPT 5.1             & 0.638 & 0.487 & 0.151 \\
Claude Sonnet 4.5   & 0.651 & 0.491 & 0.159 \\
Gemini 3 Flash       & 0.614 & 0.523 & 0.091 \\
Gemini 3.1 Pro       & 0.686 & 0.450 & 0.236 \\
\midrule
DeepSeek V3.2       & 0.620 & 0.416 & 0.204 \\
Llama 3.3 70B       & 0.627 & 0.468 & 0.159 \\
Qwen3 32B           & 0.670 & 0.520 & 0.150 \\
Llama 4 Maverick    & 0.590 & 0.471 & 0.119 \\
Baichuan M2         & 0.628 & 0.506 & 0.122 \\
Baichuan M3         & 0.661 & 0.471 & 0.190 \\
\bottomrule
\end{tabular}
\caption{Average Jaccard similarity for correctly vs.\ incorrectly predicted samples.
$\Delta J = \bar{J}_{\text{correct}} - \bar{J}_{\text{wrong}}$.
Correct predictions consistently retrieve higher-quality evidence, but wrong predictions still show substantial retrieval (J\,$>$\,0.4), indicating an evidence-reasoning gap rather than pure evidence absence.}
\label{tab:evidence_quality}
\end{table}

\section{Cochrane RoB 2.0 Decision Rules}
\label{sec:app_rob_rules}

To systematically evaluate the methodological quality of the included studies, we decompose the RoB 2.0 assessment into five standard domains. The following figures show the detailed derivation rules and logical decision maps for each domain.

\subsection{Domain 1\textnormal{:} Bias Arising from the Randomization Process}

\begin{figure}[t]
\centering
\includegraphics[width=0.9\columnwidth]{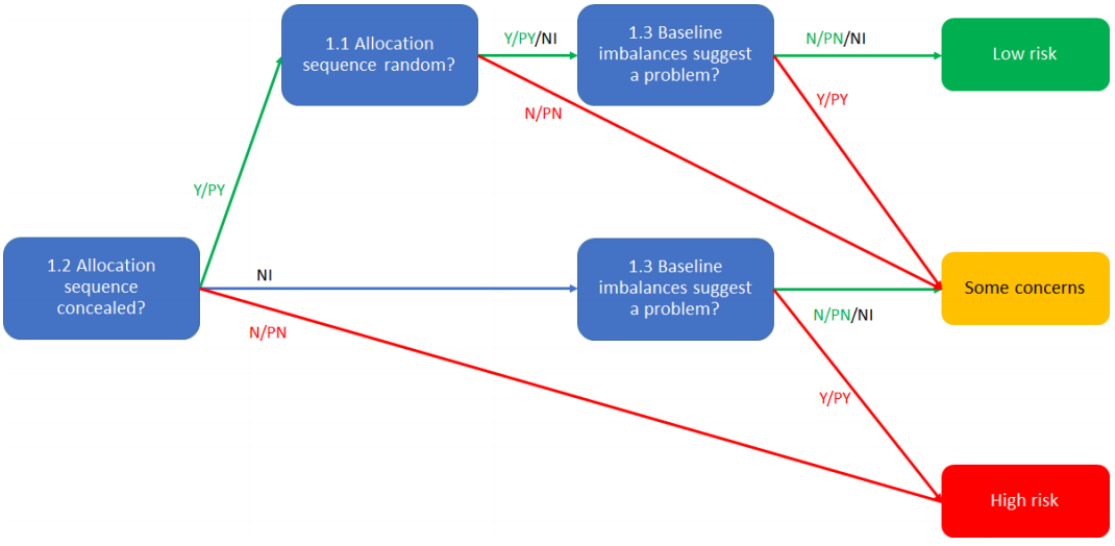}
\caption{The logical inference pathway for Domain 1 assesses bias arising from the randomization process.}
\label{fig:rob_d1}
\end{figure}

\paragraph{Rule Specification\textnormal{:}} This domain assesses whether the allocation sequence was adequately generated and concealed, and whether baseline differences suggest a problem with randomization. The specific logical flow is illustrated in Figure~\ref{fig:rob_d1}.

\subsection{Domain 2\textnormal{:} Bias Due to Deviations from Intended Interventions}

\begin{figure}[t]
\centering
\includegraphics[width=0.9\columnwidth]{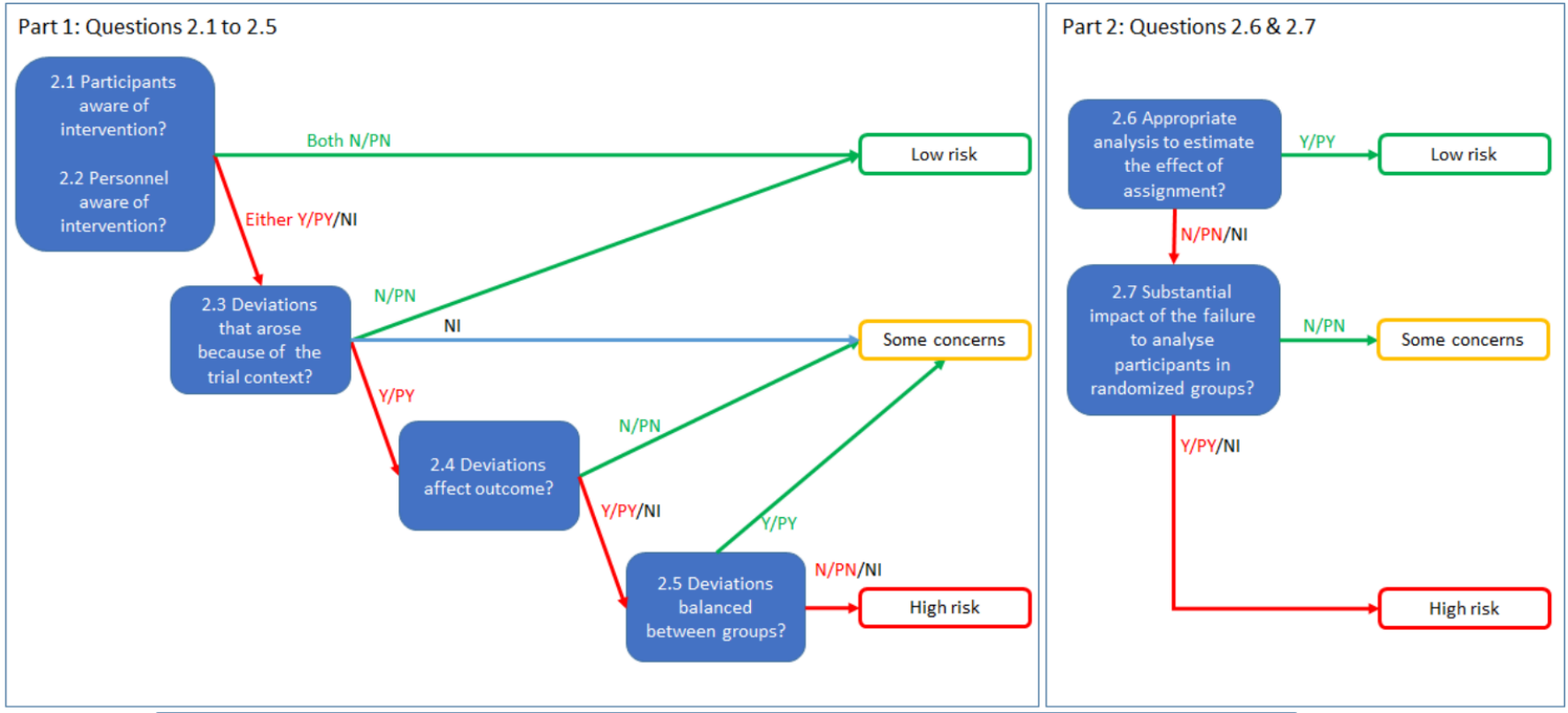}
\caption{The logical decision map for Domain 2 covers bias due to deviations from intended interventions.}
\label{fig:rob_d2}
\end{figure}

\paragraph{Rule Specification\textnormal{:}} This domain covers the rules for evaluating deviations from the intended interventions, distinguishing between the effect of assignment to an intervention and the effect of adhering to an intervention. See Figure~\ref{fig:rob_d2} for the complete logic.

\subsection{Domain 3\textnormal{:} Bias Due to Missing Outcome Data}

\begin{figure}[t]
\centering
\includegraphics[width=0.9\columnwidth]{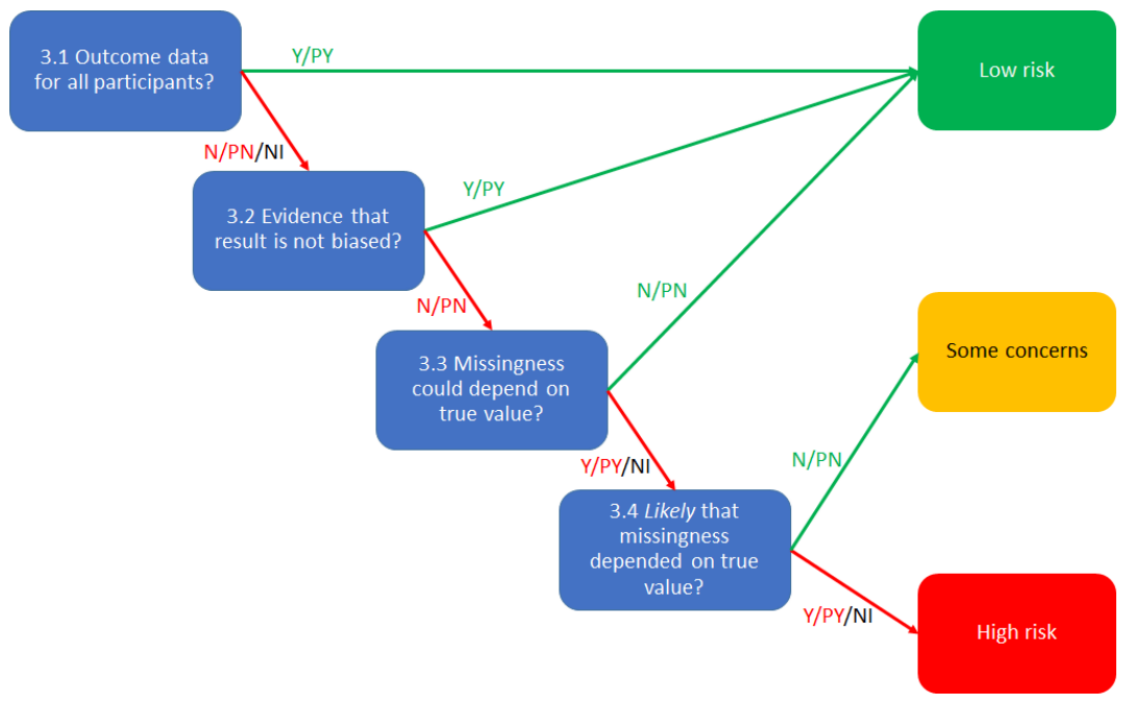}
\caption{The logical decision map for Domain 3 covers bias due to missing outcome data.}
\label{fig:rob_d3}
\end{figure}

\paragraph{Rule Specification\textnormal{:}} Here, we detail the judgment criteria for missing outcome data, including the proportions of missing data and whether the missingness depends on the true outcome value. The evaluation pathways are shown in Figure~\ref{fig:rob_d3}.

\subsection{Domain 4\textnormal{:} Bias in Measurement of the Outcome}

\begin{figure}[t]
\centering
\includegraphics[width=0.9\columnwidth]{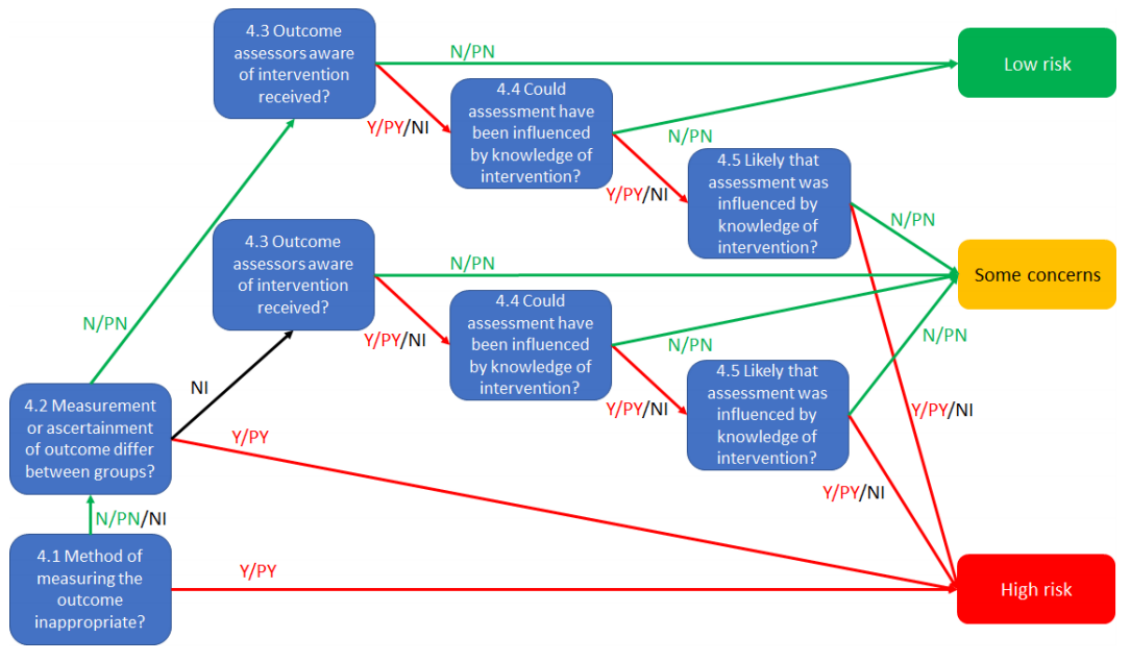}
\caption{The logical decision map for Domain 4 covers bias in measurement of the outcome.}
\label{fig:rob_d4}
\end{figure}

\paragraph{Rule Specification\textnormal{:}} This domain covers the rules concerning the appropriateness of the outcome measurement method and whether the measurement or ascertainment could differ between intervention groups. The structural rules are mapped in Figure~\ref{fig:rob_d4}.

\subsection{Domain 5\textnormal{:} Bias in Selection of the Reported Result}

\begin{figure}[t]
\centering
\includegraphics[width=0.9\columnwidth]{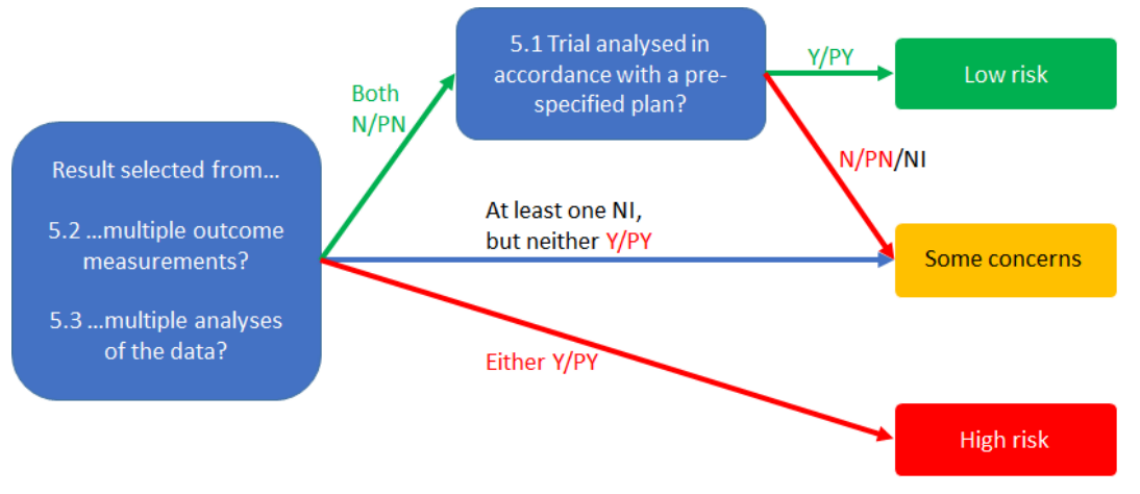}
\caption{The logical decision map for Domain 5 covers bias in selection of the reported result.}
\label{fig:rob_d5}
\end{figure}

\paragraph{Rule Specification\textnormal{:}} The final domain captures the logic for identifying selective reporting, including multiple eligible outcome measurements or multiple analyses of the data. The decision nodes are detailed in Figure~\ref{fig:rob_d5}.

\section{Cochrane RoB 1.0 Assessment Structure}
\label{sec:app_rob_rules_rob1}

\begin{figure}[t]
\centering
\includegraphics[width=0.9\columnwidth]{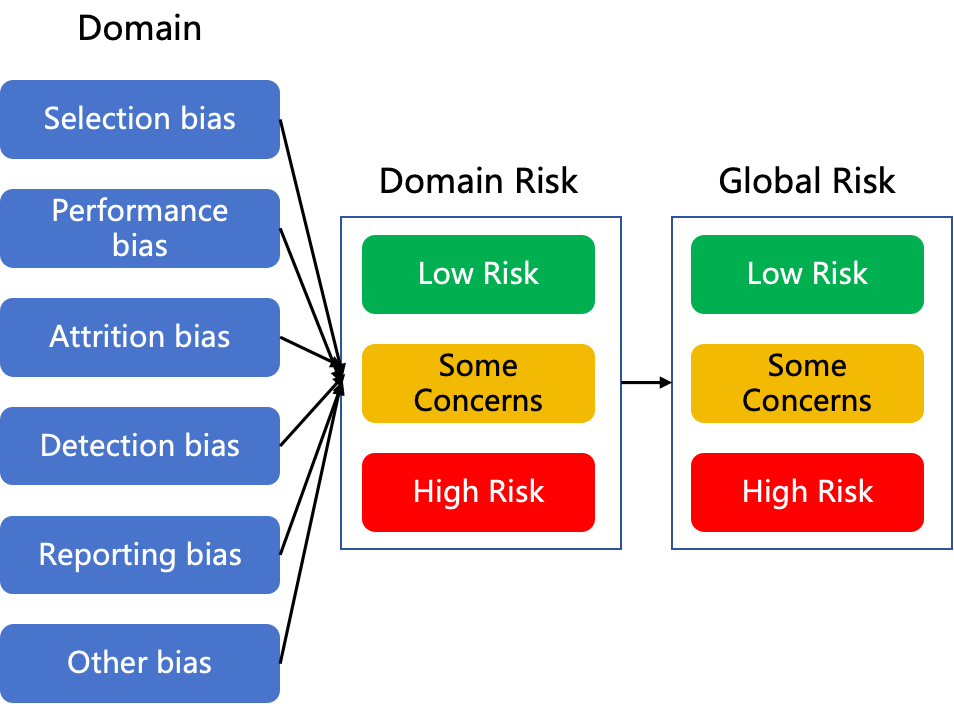}
\caption{The Cochrane Risk of Bias 1.0 assessment used in Track B assigns low, unclear, or high risk judgments across six bias categories.}
\label{fig:rob1_flow}
\end{figure}

\paragraph{The RoB 1.0 Evidence-Justification Workflow\textnormal{:}} Beyond internal reasoning, evaluating textual grounding is fundamental to clinical trustworthiness. To systematically assess methodological quality, we adopt the classic Cochrane Risk of Bias 1.0 tool, which decomposes the evaluation into six key bias categories (Figure~\ref{fig:rob1_flow}). This framework supports a separate risk judgment for each category, and our primary motivation for its inclusion is its strict evidence justification mandate. The standard explicitly requires experts to extract exact sentences to justify each rating. By adopting these human-annotated quotes as the gold standard, we audit a model's \textbf{Evidential Faithfulness}. This ensures that a predicted risk label is genuinely grounded in the correct underlying evidence rather than being a lucky ``Blind Guess''. It also verifies that conclusions stem from factual text instead of spurious correlations.

\section{Case Studies}
\label{sec:case_studies}

To provide a more intuitive understanding of the failure modes exhibited by large language models in the LogiMed-RoB evaluation, we present representative case studies across our evaluation dimensions.

\subsection{Atomic Consistency Violation}
\label{subsec:case_atomic}

For highly capable large language models, violations of Atomic Consistency are exceedingly rare. These state-of-the-art models exhibit strong instruction-following capabilities, enabling them to comprehend and follow the conditional prerequisites of each signaling question. Nevertheless, the example below illustrates a structural violation observed during the evaluation.

\begin{tcolorbox}[width=\columnwidth, 
    colback=backblue, 
    colframe=myblue, 
    title={\centering\textcolor{white}{\textbf{Track A Case Study}\textnormal{:}\textbf{ Violation of Atomic Consistency}}},
    coltitle=white, 
    fonttitle=\bfseries, 
    colbacktitle=myblue, 
    enhanced, 
    breakable, 
    pad at break*=1mm]

    \raggedright 
    \small 

    \vspace{1mm}
    \noindent\textbf{Domain}\textnormal{:} Bias due to missing outcome data \\ \textbf{Model Identifier}\textnormal{:} gemini-3.1-pro-preview\\

    \vspace{2mm}
    \noindent\textbf{Signaling Questions (Logical Prerequisites)}\textnormal{:} \\
    \textbf{3.1} Were data for this outcome available for all, or nearly all, participants randomised?\\
    \textbf{3.2} \textit{If N/PN/NI to 3.1:} Is there evidence that the result was not biased by missing outcome data?\\
    \textbf{3.3} \textit{If N/PN to 3.2:} Could missingness in the outcome depend on its true value?\\
    \textbf{3.4} \textit{If Y/PY/NI to 3.3:} Is it likely that missingness in the outcome depended on its true value?\\

    \vspace{3mm}
    \noindent\textbf{LLM Output Snippet (JSON)}\textnormal{:} \\
    \texttt{\{} \\
    \texttt{\ \ "missing\_outcome\_data": \{} \\
    \texttt{\ \ \ \ "signaling\_questions": [} \\
    \texttt{\ \ \ \ \ \ \{} \\
    \texttt{\ \ \ \ \ \ \ \ "id": "3.1",} \\
    \texttt{\ \ \ \ \ \ \ \ "answer\_code": "PN",} \\
    \texttt{\ \ \ \ \ \ \ \ "evidence\_quote": "*Calcium: N = 90. Placebo: N = 88."} \\
    \texttt{\ \ \ \ \ \ \},} \\
    \texttt{\ \ \ \ \ \ \{} \\
    \texttt{\ \ \ \ \ \ \ \ "id": "3.2",} \\
    \texttt{\ \ \ \ \ \ \ \ "answer\_code": "NI",} \\
    \texttt{\ \ \ \ \ \ \ \ "evidence\_quote": "No information provided in text"} \\
    \texttt{\ \ \ \ \ \ \},} \\
    \texttt{\ \ \ \ \ \ \{} \\
    \texttt{\ \ \ \ \ \ \ \ "id": "3.3",} \\
    \texttt{\ \ \ \ \ \ \ \ "answer\_code": "PN",} \\
    \texttt{\ \ \ \ \ \ \ \ "evidence\_quote": "Three patients in the placebo group and four in the..."} \\
    \texttt{\ \ \ \ \ \ \},} \\
    \texttt{\ \ \ \ \ \ \{} \\
    \texttt{\ \ \ \ \ \ \ \ "id": "3.4",} \\
    \texttt{\ \ \ \ \ \ \ \ "answer\_code": "NA",} \\
    \texttt{\ \ \ \ \ \ \ \ "evidence\_quote": "No information provided in text"} \\
    \texttt{\ \ \ \ \ \ \}} \\
    \texttt{\ \ \ \ ],} \\
    \texttt{\ \ \ \ "domain\_risk": "Low risk of bias"} \\
    \texttt{\ \ \}} \\
    \texttt{\}} \\

    \vspace{2mm}
    \noindent\textbf{Explanation of Logical Breakdown}\textnormal{:} \\
    As defined by the RoB 2.0 guidelines, the prerequisite for Question 3.3 is explicitly stated as ``\textit{If N/PN to 3.2}''. In this instance, the model generated an answer code of ``NI'' for Question 3.2. Consequently, the logically correct response for Question 3.3 should be ``NA'' (Not Applicable), meaning that the question should have been bypassed. By erroneously outputting an answer code of ``PN'' for Question 3.3, the model violates the established Atomic Consistency rules.
    
\end{tcolorbox}

\subsection{Domain Consistency Violation}
\label{subsec:case_domain}

A violation of Domain Consistency occurs when a model generates valid atomic answers without structural errors but fails to aggregate these answers into the correct domain-level risk label prescribed by the Cochrane decision map. This highlights a critical disconnect between information extraction and rule-based symbolic deduction. The following case illustrates this phenomenon.

\begin{tcolorbox}[width=\columnwidth, 
    colback=backblue, 
    colframe=myblue, 
    title={\centering\textcolor{white}{\textbf{Track A Case Study}\textnormal{:}\textbf{ Violation of Domain Consistency}}},
    coltitle=white, 
    fonttitle=\bfseries, 
    colbacktitle=myblue, 
    enhanced, 
    breakable, 
    pad at break*=1mm]

    \raggedright
    \small 

    \vspace{1mm}
    \noindent\textbf{Domain}\textnormal{:} Bias due to deviations from intended interventions \\ \textbf{Model Identifier}\textnormal{:} gemini-3.1-pro-preview\\

    \vspace{2mm}
    \noindent\textbf{Signaling Questions (Logical Prerequisites)}\textnormal{:} \\
    \textbf{2.1} Were participants aware of their assigned intervention during the trial?\\
    \textbf{2.2} Were carers and people delivering the interventions aware of participants' assigned intervention during the trial?\\
    \textbf{2.3} \textit{If Y/PY/NI to 2.1 or 2.2:} Were there deviations from the intended intervention that arose because of the trial context?\\
    \textbf{2.4} \textit{If Y/PY/NI to 2.3:} Were these deviations likely to have affected the outcome?\\
    \textbf{2.5} \textit{If Y/PY to 2.4:} Were these deviations from intended intervention balanced between groups?\\
    \textbf{2.6} Was an appropriate analysis used to estimate the effect of assignment to intervention?\\
    \textbf{2.7} \textit{If N/PN/NI to 2.6:} Was there potential for a substantial impact (on the result) of the failure to analyse participants in the group to which they were randomised?\\

    \vspace{3mm}
    \noindent\textbf{LLM Output Snippet (JSON)}\textnormal{:} \\
    \texttt{\{} \\
    \texttt{\ \ "intended\_interventions": \{} \\
    \texttt{\ \ \ \ "signaling\_questions": [} \\
    \texttt{\ \ \ \ \ \ \{} \\
    \texttt{\ \ \ \ \ \ \ \ "id": "2.1", "answer\_code": "Y",} \\
    \texttt{\ \ \ \ \ \ \ \ "evidence\_quote": "Participants received either TEAM (n = 19) or TAU (n = 19)."} \\
    \texttt{\ \ \ \ \ \ \},} \\
    \texttt{\ \ \ \ \ \ \{} \\
    \texttt{\ \ \ \ \ \ \ \ "id": "2.2", "answer\_code": "Y",} \\
    \texttt{\ \ \ \ \ \ \ \ "evidence\_quote": "The TEAM intervention was delivered collaboratively by a nurse..."} \\
    \texttt{\ \ \ \ \ \ \},} \\
    \texttt{\ \ \ \ \ \ \{} \\
    \texttt{\ \ \ \ \ \ \ \ "id": "2.3", "answer\_code": "N",} \\
    \texttt{\ \ \ \ \ \ \ \ "evidence\_quote": "Individuals in TAU continued treatment with their regular medical..."} \\
    \texttt{\ \ \ \ \ \ \},} \\
    \texttt{\ \ \ \ \ \ \{} \\
    \texttt{\ \ \ \ \ \ \ \ "id": "2.4", "answer\_code": "NA",} \\
    \texttt{\ \ \ \ \ \ \ \ "evidence\_quote": "No information provided in text"} \\
    \texttt{\ \ \ \ \ \ \},} \\
    \texttt{\ \ \ \ \ \ \{} \\
    \texttt{\ \ \ \ \ \ \ \ "id": "2.5", "answer\_code": "NA",} \\
    \texttt{\ \ \ \ \ \ \ \ "evidence\_quote": "No information provided in text"} \\
    \texttt{\ \ \ \ \ \ \},} \\
    \texttt{\ \ \ \ \ \ \{} \\
    \texttt{\ \ \ \ \ \ \ \ "id": "2.6", "answer\_code": "Y",} \\
    \texttt{\ \ \ \ \ \ \ \ "evidence\_quote": "Note that analyses were conducted on participants with complete data."} \\
    \texttt{\ \ \ \ \ \ \},} \\
    \texttt{\ \ \ \ \ \ \{} \\
    \texttt{\ \ \ \ \ \ \ \ "id": "2.7", "answer\_code": "NA",} \\
    \texttt{\ \ \ \ \ \ \ \ "evidence\_quote": "No information provided in text"} \\
    \texttt{\ \ \ \ \ \ \}} \\
    \texttt{\ \ \ \ ],} \\
    \texttt{\ \ \ \ "domain\_risk": "Low risk of bias"} \\
    \texttt{\ \ \}} \\
    \texttt{\}} \\

    \vspace{2mm}
    \noindent\textbf{Explanation of Logical Breakdown}\textnormal{:} \\
    According to the established RoB 2.0 decision knowledge graph, the specific combination of atomic answers extracted by the model (\textbf{2.1: Y}, \textbf{2.2: Y}, \textbf{2.3: N}, and \textbf{2.6: Y}) deterministically maps to a domain risk of ``\textbf{Some concerns}''. However, the model incorrectly outputs ``\textbf{Low risk of bias}''. This discrepancy demonstrates a direct violation of Domain Consistency because the model's domain risk judgment contradicts the output prescribed by the expert-defined rules.
    
\end{tcolorbox}

\subsection{Aggregation Consistency Violation}
\label{subsec:case_aggregation}

When computing the overall risk by aggregating the risk judgments of all individual domains, explicit errors are relatively rare for highly capable models. This is because state-of-the-art large language models have generally internalized the Worst-of Principle required for final RoB decisions during pretraining. When an aggregation error does occur, it reflects a failure to apply the Worst-of Principle and may be compounded by incorrect domain-level assessments. The following model output illustrates this process.

\begin{tcolorbox}[width=\columnwidth, 
    colback=backblue, 
    colframe=myblue, 
    title={\centering\textcolor{white}{\textbf{Track A Case Study}\textnormal{:}\textbf{ Violation of Aggregation Consistency}}},
    coltitle=white, 
    fonttitle=\bfseries, 
    colbacktitle=myblue, 
    enhanced, 
    breakable, 
    pad at break*=1mm]

    \raggedright
    \small 

    \vspace{1mm}
    \noindent\textbf{Evaluation Phase}\textnormal{:} Global Risk Aggregation \\
    \textbf{Model Identifier}\textnormal{:} gemini-3.1-pro-preview\\

    \vspace{2mm}
    \noindent\textbf{LLM Output vs.\ Ground Truth (JSON Format)}\textnormal{:} \\
    \texttt{\{} \\
    \texttt{\ \ "llm\_result": \{} \\
    \texttt{\ \ \ \ "randomisation\_process": \{"domain\_risk": "Some concerns"\},} \\
    \texttt{\ \ \ \ "intended\_interventions": \{"domain\_risk": "Some concerns"\},} \\
    \texttt{\ \ \ \ "missing\_outcome\_data": \{"domain\_risk": "Low risk of bias"\},} \\
    \texttt{\ \ \ \ "measurement\_outcome": \{"domain\_risk": "Some concerns"\},} \\
    \texttt{\ \ \ \ "selection\_reported\_result": \{"domain\_risk": "Some concerns"\},} \\ 
    \texttt{\ \ \ \ "overall\_risk": "High risk of bias"} \\
    \texttt{\ \ \},} \\
    \texttt{\ \ "ground\_truth": \{} \\
    \texttt{\ \ \ \ "randomisation\_process\_judgment": "Some concerns",} \\
    \texttt{\ \ \ \ "intended\_interventions\_judgment": "Low risk of bias",} \\
    \texttt{\ \ \ \ "missing\_outcome\_data\_judgment": "Low risk of bias",} \\
    \texttt{\ \ \ \ "measurement\_outcome\_judgment": "Low risk of bias",} \\
    \texttt{\ \ \ \ "selection\_reported\_result\_judgment": "Some concerns",} \\
    \texttt{\ \ \ \ "overall\_risk": "Some concerns"} \\
    \texttt{\ \ \}} \\
    \texttt{\}} \\

    \vspace{2mm}
    \noindent\textbf{Explanation of Logical Breakdown}\textnormal{:} \\
    As observed in the output, the large language model incorrectly assessed the risks for two domains, misclassifying intended interventions and measurement outcomes as ``Some concerns''. More critically, the model exhibits a severe violation of internal Aggregation Consistency: its own extracted domain risks consist entirely of ``Low risk'' and ``Some concerns'', meaning that the worst-case domain risk is ``Some concerns''. However, the model outputs an ``overall\_risk'' of ``High risk of bias''. This demonstrates that the model made incorrect domain-level judgments compared with the ground truth and failed to apply the algorithmic Worst-of Principle to its own internal reasoning chain.
    
\end{tcolorbox}

\subsection{Evidential Faithfulness\textnormal{:} Blind Guess}
\label{subsec:case_blind_guess}

A ``Blind Guess'' occurs when a model's Retrieval Strength falls below the predefined threshold ($\tau$) for Evidential Faithfulness, yet it still predicts the correct final risk label. This phenomenon indicates that the model did not use the relevant textual evidence to reach its prediction. To differentiate strong from weak retrieval, we set the Retrieval Strength threshold to $\tau = 0.8$. The following example illustrates this behavior.

\begin{tcolorbox}[width=\columnwidth, 
    colback=backblue, 
    colframe=myblue, 
    title={\centering\textcolor{white}{\textbf{Track B Case Study}\textnormal{:}\textbf{ Evidential Faithfulness (Blind Guess)}}},
    coltitle=white, 
    fonttitle=\bfseries, 
    colbacktitle=myblue, 
    enhanced, 
    breakable, 
    pad at break*=1mm]

    \raggedright
    \small 

    \vspace{1mm}
    \noindent\textbf{Domain}\textnormal{:} Incomplete outcome data (attrition bias) \\
    \textbf{Model Identifier}\textnormal{:} gemini-3.1-pro-preview\\

    \vspace{2mm}
    \noindent\textbf{Context Snippet (Indexed Sentences)}\textnormal{:} \\
    \textbf{[0]} the smaller the non-inferiority margin, the greater the sample size, so to set the margin closer to the expected 25\% difference in effectiveness would have required an unrealistically large sample size.\\
    \textbf{[1]} therefore, the non-inferiority margin was set at 37\% (the 25\% expected difference plus a further 12\%) which meant that 234 participants would be sufficient.\\
    \textbf{[2]} thus 37\% was the maximum allowable difference for a strategy of starting treatment with doxycycline to be considered non-inferior.\\
    \textbf{[3]} given these considerations and the sample size required for the safety analysis it was decided to recruit a total of 256 patients.\\
    \textbf{[4]} for initial treatment with doxycycline to be considered an acceptable alternative strategy to prednisolone, non-inferiority had to be shown for effectiveness and superiority for safety.\\
    \textbf{[5]} analysis used a binomial regression model with an identity link function adjusting for baseline severity of bullous pemphigoid, patient age, and karnofsky score for functional impairment to estimate the absolute difference between the two treatment groups, and missing data were imputed.\\
    \textbf{[6]} superiority analyses were done on a modified intention-to-treat basis (participants who fulfilled eligibility were randomly assigned to either of the study drugs and had data on the outcome of interest as pre-defined in our protocol) and non-inferiority analyses were done on both modified intention-to-treat and per-protocol populations according to recommended practice23 and according to the protocol and statistical analysis plan.\\

    \vspace{3mm}
    \noindent\textbf{Prompt Task}\textnormal{:} \\
    You must complete \textbf{two tasks}:\\
    1. \textbf{Identify Evidence Sentences}\textnormal{:} Determine which sentence indices (0-based) from the context provide accurate evidence. Return the indices as an array of integers.\\
    2. \textbf{Determine Risk of Bias}\textnormal{:} Based on the evidence sentences you identified, determine the risk of bias judgment.\\

    \vspace{3mm}
    \noindent\textbf{LLM Output Snippet (JSON)}\textnormal{:} \\
    \texttt{\{} \\
    \texttt{\ \ "evidence\_indices": [5, 6],} \\
    \texttt{\ \ "risk\_of\_bias": "low"} \\
    \texttt{\}} \\

    \vspace{3mm}
    \noindent\textbf{Ground Truth}\textnormal{:} \\
    \texttt{\{} \\
    \texttt{\ \ "evidence\_indices": [3],} \\
    \texttt{\ \ "risk\_of\_bias": "low"} \\
    \texttt{\}} \\

    \vspace{2mm}
    \noindent\textbf{Explanation of Evidential Failure}\textnormal{:} \\
    The ground truth indicates that the correct risk label is ``low'', supported exclusively by sentence \textbf{[3]}. The model successfully predicted the ``low'' risk label. However, it retrieved sentences \textbf{[5]} and \textbf{[6]} as its evidence and missed the actual justification. Consequently, the Jaccard similarity between the retrieved evidence and the gold standard is $0$ (which is strictly $< \tau$). This demonstrates that the model's prediction was not grounded in the correct textual evidence.
    
\end{tcolorbox}

\subsection{Evidential Faithfulness\textnormal{:} Reasoning Failure}
\label{subsec:case_reasoning_failure}

Unlike a ``Blind Guess'', a ``Reasoning Failure'' occurs when a model successfully identifies and retrieves the correct textual evidence (\ie Retrieval Strength $\ge \tau$) but fails to deduce the correct risk label from that evidence. This phenomenon isolates a specific deficiency in the model's rule-based integration and clinical reasoning capabilities. Although its information extraction is accurate, its decision-making logic remains misaligned with expert consensus. The following case illustrates this limitation.

\begin{tcolorbox}[width=\columnwidth, 
    colback=backblue, 
    colframe=myblue, 
    title={\centering\textcolor{white}{\textbf{Track B Case Study}\textnormal{:}\textbf{ Evidential Faithfulness (Reasoning Failure)}}},
    coltitle=white, 
    fonttitle=\bfseries, 
    colbacktitle=myblue, 
    enhanced, 
    breakable, 
    pad at break*=1mm]

    \raggedright
    \small 

    \vspace{1mm}
    \noindent\textbf{Domain}\textnormal{:} Blinding of participants and personnel (performance bias)\\
    \textbf{Model Identifier}\textnormal{:} gemini-3.1-pro-preview\\

    \vspace{2mm}
    \noindent\textbf{Context Snippet (Indexed Sentences)}\textnormal{:} \\
    \textbf{[0]} some limitations of our study should be recognized.\\
    \textbf{[1]} although we demonstrated that vildagliptin improves epc bioavailability, the underlying pathophysiological explanation remains unclear and future studies are warranted to unravel sdf-1$\alpha$ dependent and independent mechanisms.\\
    \textbf{[2]} the clinical impact of the epc and sdf-1$\alpha$ changes induced by vildagliptin, although concordantly pointing to a potential beneficial effect, remains unknown.\\
    \textbf{[3]} the open-label nature of the study also needs to be acknowledged.\\
    \textbf{[4]} in conclusion, vildagliptin exerts a beneficial long-term effect on circulating epc levels, at glucose equipoise, with a putative positive effect on vascular integrity.\\
    \textbf{[5]} the vildagliptin-induced reduction in plasma sdf-1$\alpha$ levels might be desirable in light of the emerging role of circulating sdf-1$\alpha$ as an independent cardiovascular risk biomarker.\\
    \textbf{[6]} abbreviations apcallophycocyanin  bnpbrain natriuretic peptide  ckdchronic kidney disease  crpc-reactive protein  cvcardiovascular  dpp-4idipeptidyl peptidase-4 inhibitors  epcendothelial progenitor cells  fitcfluorescein isothiocyanate  fpgfasting plasma glucose  glmgeneral linear model  glp-1glucagon-like peptide-1  ittintention to treat  kdrkinase insert domain receptor  mabmonoclonal antibody  pephycoerythrin  percpperidinin chlorophyll protein complex  riccardo c.\\
    \textbf{[7]} bonadonna and ivana zavaroni jointly supervised this work  authors’ contributions adc wrote the manuscript.\\

    \vspace{3mm}
    \noindent\textbf{Prompt Task}\textnormal{:} \\
    You must complete \textbf{two tasks}:\\
    1. \textbf{Identify Evidence Sentences}\textnormal{:} Determine which sentence indices (0-based) from the context provide accurate evidence. Return the indices as an array of integers.\\
    2. \textbf{Determine Risk of Bias}\textnormal{:} Based on the evidence sentences you identified, determine the risk of bias judgment.\\

    \vspace{3mm}
    \noindent\textbf{LLM Output Snippet (JSON)}\textnormal{:} \\
    \texttt{\{} \\
    \texttt{\ \ "evidence\_indices": [3],} \\
    \texttt{\ \ "risk\_of\_bias": "high"} \\
    \texttt{\}} \\

    \vspace{3mm}
    \noindent\textbf{Ground Truth}\textnormal{:} \\
    \texttt{\{} \\
    \texttt{\ \ "evidence\_indices": [3],} \\
    \texttt{\ \ "risk\_of\_bias": "low"} \\
    \texttt{\}} \\

    \vspace{2mm}
    \noindent\textbf{Explanation of Evidential Failure}\textnormal{:} \\
    In this instance, the model accurately pinpointed sentence \textbf{[3]} (``the open-label nature of the study also needs to be acknowledged.'') as the critical piece of evidence. However, despite retrieving the same justification as the human experts, the model erroneously concluded that the risk of bias was ``high''. In the specific clinical context evaluated by the experts, this methodological characteristic was judged to have a ``low'' risk. This discrepancy shows that the model failed to interpret the clinical nuance of the retrieved fact correctly.
    
\end{tcolorbox}


\section{Output Format Compliance and Validation}
\label{sec:app_format_compliance}

A potential concern regarding the near-zero Complete Consistency Rates observed for certain models (\eg Llama 3.3 70B at 0.00\% atomic, Qwen3 32B and Llama 4 Maverick at 1--2\% E2E) is whether these results reflect genuine reasoning failures or mere formatting noncompliance. We address this concern with two lines of evidence.

\subsection{Robust Parsing Pipeline with Automatic Retry}

Our evaluation pipeline enforces a strict output contract: every model response must be valid JSON containing the exact required answer codes from the predefined answer space $\mathcal{A} = \{\text{Y, PY, PN, N, NI, NA}\}$. If a response fails JSON parsing or contains answer codes outside the valid set, the instance is automatically routed to an error queue and retried. Only successfully parsed, structurally valid outputs contribute to the reported metrics.

Table~\ref{tab:format_compliance} reports the number of successfully parsed trials per model. The model-specific counts range from 554 to 626 valid trials. All reported consistency metrics use only these successfully parsed outputs, so malformed responses are excluded rather than scored as logical failures.

\begin{table}[tbp]
\centering
\small
\begin{tabular}{lc}
\toprule
\textbf{Model} & \textbf{Successfully Parsed Trials} \\
\midrule
\multicolumn{2}{l}{\textit{Proprietary}} \\
GPT 5.1 & 626 \\
Claude Sonnet 4.5 & 618 \\
Gemini 3 Flash & 625 \\
Gemini 3.1 Pro & 554 \\
\midrule
\multicolumn{2}{l}{\textit{Open-weight}} \\
DeepSeek V3.2 & 626 \\
Llama 3.3 70B & 626 \\
Qwen3 32B & 625 \\
Llama 4 Maverick & 620 \\
Baichuan M2 & 625 \\
Baichuan M3 & 570 \\
\bottomrule
\end{tabular}
\caption{Number of successfully parsed trials per model in the total evaluation set. Reported metrics are calculated only from these structurally valid outputs.}
\label{tab:format_compliance}
\end{table}

\subsection{Theoretical Validation of Error Compounding}

Beyond empirical parsing validation, the near-zero Complete Consistency Rates are consistent with the structure of our evaluation. Each trial assessment contains 22 signaling questions, of which 10 prerequisite-constrained questions in D2--D4 enter the CAR calculation. Complete atomic consistency requires all 10 branch constraints to be satisfied for a trial.

Under an independence and equal-accuracy approximation, the expected complete atomic consistency can be estimated from the per-question CAR reported in Table~\ref{tab:main_results}:
\begin{equation}
P(\text{Complete Atomic}) \approx (\text{CAR}_{\text{overall}})^{10}.
\end{equation}

For example:
\begin{itemize}[leftmargin=*]
    \item Llama 3.3 70B: $\text{CAR} = 68.18\% \Rightarrow (0.6818)^{10} \approx 2.17\%$
    \item Qwen3 32B: $\text{CAR} = 80.83\% \Rightarrow (0.8083)^{10} \approx 11.90\%$
    \item Gemini 3.1 Pro: $\text{CAR} = 98.88\% \Rightarrow (0.9888)^{10} \approx 89.35\%$
\end{itemize}

The empirically observed values are 0.00\%, 8.80\%, and 88.99\%, respectively. The approximation is close for the two stronger models, while deviations are expected because branch errors are correlated and the number of valid trials varies by model. The calculation supports error compounding as an explanation without treating independence as an exact model of the data.

\subsection{Summary}

The Complete Consistency Rate was deliberately designed as a stringent end-to-end measure to quantify whether LLMs can reliably navigate all conditional branches in a complex medical decision tree without succumbing to error compounding. The agreement between the approximation and the stronger models' empirical results supports the interpretation that the reported scores reflect logical reasoning behavior rather than only formatting artifacts.

\section{Prompt Templates}
\label{app:prompt_templates}

In this section, we provide the complete set of prompt templates used throughout our experiments. The exact wording for each task is presented in full below.

\onecolumn

\begin{tcolorbox}[
    width=\linewidth,
    colback=backblue,
    colframe=gray,
    title={\textcolor{white}{\textbf{Domain-General Prompt Template}}},
    coltitle=white,
    fonttitle=\bfseries,
    colbacktitle=gray,
    enhanced,
    breakable,
    fontupper=\normalsize
]

\textbf{\# Role: Clinical Trial Methodologist (RoB 2 Specialist)}

You are an expert in critical appraisal of clinical trials using the Cochrane Risk of Bias 2 (RoB 2) tool.
Your task is to determine the overall risk of bias judgment for the entire trial based on the five domain-level risk assessments.

\vspace{0.5em}
\textbf{\# Target Outcome}

Please assess the overall risk of bias for this specific outcome: \texttt{\{outcome\}}

\vspace{0.5em}
\textbf{\# Domain-Level Risk Assessments}

The following are the risk of bias assessments for each of the five RoB 2 domains:

\vspace{0.5em}
\textbf{1. Bias arising from the randomisation process}\\
\textbf{Domain Risk}: \texttt{\{randomisation\_process\_risk\}}

\vspace{0.5em}
\textbf{2. Bias due to deviations from intended interventions}\\
\textbf{Domain Risk}: \texttt{\{intended\_interventions\_risk\}}

\vspace{0.5em}
\textbf{3. Bias due to missing outcome data}\\
\textbf{Domain Risk}: \texttt{\{missing\_outcome\_data\_risk\}}

\vspace{0.5em}
\textbf{4. Bias in measurement of the outcome}\\
\textbf{Domain Risk}: \texttt{\{measurement\_outcome\_risk\}}

\vspace{0.5em}
\textbf{5. Bias in selection of the reported result}\\
\textbf{Domain Risk}: \texttt{\{selection\_reported\_result\_risk\}}

\vspace{0.5em}
\textbf{\# Task}

Based on the five domain-level risk assessments above, determine the \textbf{overall risk of bias} for this trial using your expert knowledge of RoB 2 guidance.

\vspace{0.5em}
\textbf{\# Return format (JSON Only)}

You must output a strictly valid JSON object. Do not wrap the JSON in markdown code blocks (like \texttt{```json ... ```}). Output raw JSON only.

Example:\\
\texttt{\{\\
\hspace*{1em}"overall\_risk": "Low risk of bias / Some concerns / High risk of bias",\\
\hspace*{1em}"reasoning": "Brief explanation of your judgment (optional)"\\
\}}

Allowed values for \texttt{overall\_risk}:
\begin{itemize}[leftmargin=*]
    \item \textbf{Low risk of bias}
    \item \textbf{Some concerns}
    \item \textbf{High risk of bias}
\end{itemize}

\end{tcolorbox}

\begin{tcolorbox}[
    width=\linewidth,
    colback=backblue,
    colframe=gray,
    title={\textcolor{white}{\textbf{Overall Risk Assessment Prompt}}},
    coltitle=white,
    fonttitle=\bfseries,
    colbacktitle=gray,
    enhanced,
    breakable,
    fontupper=\normalsize
]

\textbf{\# Role: Risk of Bias (RoB) Expert}

You are an expert in critical appraisal of clinical trials using the Cochrane Risk of Bias assessment tool.
Your task is to identify evidence sentences that accurately answer the given question and determine the risk of bias judgment.

\vspace{0.5em}
\textbf{\# Bias Type}

\texttt{\{bias\}}

\vspace{0.5em}
\textbf{\# Question}

\texttt{\{question\}}

\vspace{0.5em}
\textbf{\# Context Sentences}

The following is a list of sentences extracted from the study manuscript. Each sentence is numbered starting from 0.

\texttt{\{context\_list\}}

\vspace{0.5em}
\textbf{\# Task}

You must complete \textbf{two tasks}:

\begin{enumerate}[leftmargin=*]
    \item \textbf{Identify Evidence Sentences}: Determine which sentence indices (0-based) from the \texttt{context\_list} provide accurate evidence to answer the question. You may select multiple sentences if they are all relevant. Return the indices as an array of integers.
    \item \textbf{Determine Risk of Bias}: Based on the evidence sentences you identified, determine the risk of bias judgment according to the criteria provided in the question.
\end{enumerate}

\vspace{0.5em}
\textbf{\# Risk of Bias Judgment}

Allowed values for \texttt{risk\_of\_bias}:
\begin{itemize}[leftmargin=*]
    \item \textbf{low} (Low risk of bias)
    \item \textbf{some concerns} (Some concerns)
    \item \textbf{high} (High risk of bias)
\end{itemize}

\vspace{0.5em}
\textbf{\# Return format (JSON Only)}

You must output a strictly valid JSON object. Do not wrap the JSON in markdown code blocks (like \texttt{```json ... ```}). Output raw JSON only.

Example:\\
\texttt{\{\\
\hspace*{1em}"evidence\_indices": [0, 2, 5],\\
\hspace*{1em}"risk\_of\_bias": "low"\\
\}}

\vspace{0.5em}
\textbf{\# Important Notes}

\textbf{1. Evidence Indices}:
\begin{itemize}[leftmargin=*]
    \item The indices must be 0-based (the first sentence is index 0, the second is index 1, etc.)
    \item You may select multiple sentences if they all provide relevant evidence
    \item If no sentences provide relevant evidence, return an empty array: []
    \item Only include indices that directly support answering the question
\end{itemize}

\vspace{0.5em}
\textbf{2. Risk of Bias Judgment}:
\begin{itemize}[leftmargin=*]
    \item Carefully review the criteria in the question for ``low risk'', ``high risk'', and ``unclear risk''
    \item Base your judgment solely on the evidence sentences you identified
    \item Use the exact values: ``low'', ``some concerns'', or ``high''
\end{itemize}

\vspace{0.5em}
\textbf{3. Output Format}:
\begin{itemize}[leftmargin=*]
    \item Must be valid JSON
    \item Do not include any explanatory text outside the JSON
    \item The JSON object must contain exactly two fields: ``evidence\_indices'' (array of integers) and ``risk\_of\_bias'' (string)
\end{itemize}

\end{tcolorbox}

\begin{tcolorbox}[
    width=\linewidth,
    colback=backblue,
    colframe=gray,
    title={\textcolor{white}{\textbf{Randomization Process Prompt}}},
    coltitle=white,
    fonttitle=\bfseries,
    colbacktitle=gray,
    enhanced,
    breakable,
    fontupper=\normalsize
]

\textbf{1.1~Was the allocation sequence random?}

\begin{itemize}[leftmargin=*]
    \item \textbf{Y}: A randomization method is described in the sequence generation process. Examples include: use of computer-generated random numbers; random number tables; coin flipping; shuffling or envelope methods; dice rolling; lottery drawing. The ``minimization method'' is also a randomization-based grouping approach and is classified as random allocation.
    \item \textbf{N}: No random grouping method was used in the group assignment process, or group assignment was predictable. Examples include: selective enrollment; selection methods based on dates (\eg date of birth, date of visit, visit sequence) or medical record numbers; group assignment determined by physicians or patients themselves; grouping based on the accessibility of the intervention; or other biased grouping methods.
    \item \textbf{NI}: The original text only states ``random'' but does not further describe the randomization method.
    \item \textbf{PY/PN}: For a large-scale study, if it is conducted by an independent center or designed for regulatory purposes, the allocation process may be considered random (PY). If other concurrent trials by the same research team explicitly did not use random allocation, the current trial may be considered non-randomized (PN).
\end{itemize}

\vspace{0.5em}
\textbf{1.2~Was the allocation sequence concealed until participants were enrolled and assigned to interventions?}

\begin{itemize}[leftmargin=*]
    \item \textbf{Y}: Interventions are allocated using remote or centralized management methods, and the allocation process is conducted by third-party institutions independent of enrollment implementers (\eg independent pharmaceutical companies, service providers of telephone or online randomization processes). Envelopes and medication packaging are appropriately used. Envelopes should be sequentially coded, required to be opaque and hermetically sealed, with tamper-evident seals. Medication packaging should also be sequentially coded and identical in appearance. If this section is not elaborated in detail in the original text, it may be judged as PY or PN at discretion.
    \item \textbf{N}: Enrollment implementers or study participants could predict the group assignment.
\end{itemize}

\vspace{0.5em}
\textbf{1.3~Did baseline differences between intervention groups suggest a problem with the randomisation process?}

\begin{itemize}[leftmargin=*]
    \item Note that differences that are compatible with chance do not lead to a risk of bias. A small number of differences identified as ``statistically significant'' at the conventional 0.05 threshold should usually be considered to be compatible with chance.
    \item \textbf{N}: Group differences caused by random error do not introduce bias; either baseline balance exists between groups or group differences are attributable to random error (groups are comparable).
    \item \textbf{Y}: (1) substantial differences between intervention group sizes, compared with the intended allocation ratio; or (2) a substantial excess in statistically significant differences in baseline characteristics between intervention groups, beyond that expected by chance; or (3) imbalance in one or more key prognostic factors, or baseline measures of outcome variables, that is very unlikely to be due to chance and for which the between-group difference is big enough to result in bias in the intervention effect estimate. Also answer ``Yes'' if there are other reasons to suspect that the randomization process was problematic: (4) excessive similarity in baseline characteristics that is not compatible with chance.
    \item \textbf{NI}: No valid baseline information is stated in the original text. The answer to this question shall not affect the answers to ``Were study participants randomly allocated?'' and ``Was allocation concealment implemented before participant enrollment?''. For example: If a trial has significant baseline imbalance between groups, but the authors report using an appropriate randomization method, the answers to ``Were study participants randomly allocated?'' and ``Was allocation concealment implemented before participant enrollment?'' shall be judged based on the randomization method reported by the authors. All issues related to intergroup balance are evaluated in this question and reflected in the overall assessment of randomization process bias. Researchers may control for baseline imbalance between groups through statistical analysis methods to remedy problems arising from the randomization process.
    \item The answer to this question should not influence answers to questions 1.1 or 1.2. For example, if the trial has large baseline imbalances, but authors report adequate randomization methods, questions 1.1 and 1.2 should still be answered on the basis of the reported adequate methods, and any concerns about the imbalance should be raised in the answer to the question 1.3 and reflected in the domain-level risk-of-bias judgement.
    \item Trialists may undertake analyses that attempt to deal with flawed randomization by controlling for imbalances in prognostic factors at baseline. To remove the risk of bias caused by problems in the randomization process, it would be necessary to know, and measure, all the prognostic factors that were imbalanced at baseline. It is unlikely that all important prognostic factors are known and measured, so such analyses will at best reduce the risk of bias. If review authors wish to assess the risk of bias in a trial that controlled for baseline imbalances in order to mitigate failures of randomization, the study should be assessed using the ROBINS-I tool.
\end{itemize}

\end{tcolorbox}

\begin{tcolorbox}[
    width=\linewidth,
    colback=backblue,
    colframe=gray,
    title={\textcolor{white}{\textbf{Deviations from Intended Interventions Questions}}},
    coltitle=white,
    fonttitle=\bfseries,
    colbacktitle=gray,
    enhanced,
    breakable,
    fontupper=\normalsize
]

\textbf{2.1~Were participants aware of their assigned intervention during the trial?}

\begin{itemize}[leftmargin=*]
    \item \textbf{N}: Study participants were blinded (\eg with a placebo).
    \item \textbf{Y/PY}: Study participants were able to determine which intervention group they belonged to through adverse reactions or other means.
\end{itemize}

\vspace{0.5em}
\textbf{2.2~Were carers and people delivering the interventions aware of participants' assigned intervention during the trial?}

\begin{itemize}[leftmargin=*]
    \item \textbf{N}: Healthcare providers were blinded (\eg with a placebo).
    \item \textbf{Y/PY}: Healthcare providers were able to determine which intervention group study participants belonged to through adverse reactions or other means; if the randomization protocol lacked allocation concealment, healthcare providers were highly likely to be aware of the participants’ group assignment.
\end{itemize}

\vspace{0.5em}
\textbf{2.3~If Y/PY/NI to 2.1 or 2.2: Were there deviations from the intended intervention that arose because of the trial context?}

\begin{itemize}[leftmargin=*]
    \item \textbf{NA}: If not Y/PY/NI to 2.1 or 2.2.
    \item \textbf{Y}: Study participants perceived being assigned to the control group as 'unfortunate' and thus sought the same intervention as the intervention group or other interventions. Because participants had different expectations for the intervention group and control group, and the bias caused by these differing expectations is not part of routine medical practice, the evaluation of intervention effects fails to reflect the true effect of the intervention in practical settings.
    \item \textbf{N/PN}: Deviations occurring in the trial also occur in routine medical practice. Examples include: (1) discontinuation of medication due to adverse drug reactions; (2) non-adherence to the intervention; (3) 'concomitant interventions' implemented in response to intervention outcomes. Blinding cannot be implemented in trials where certain interventions have specific adverse reactions. In such cases, this item is judged as 'N' or 'PN' unless the deviation in the intervention method is related to the trial content. Termination of the intervention or crossing over to another group caused by adverse reactions is generally not considered a deviation from the prespecified intervention.
    \item \textbf{NI / PY}: If researchers do not report whether the deviation is related to the trial content, the answer should be 'NI'; however, if it can be judged that deviations related to the trial content are highly likely to have occurred, the answer should be 'PY'.
\end{itemize}

\vspace{0.5em}
\textbf{2.4~If Y/PY/NI to 2.3: Were these deviations likely to have affected the outcome?}

\begin{itemize}[leftmargin=*]
    \item \textbf{NA}: If not Y/PY/NI to 2.3.
    \item \textbf{Y/PY/N/PN}: If the deviation from the predetermined intervention is unrelated to routine medical practice, and the deviation between groups affects the outcome, this should be taken seriously.
\end{itemize}

\vspace{0.5em}
\textbf{2.5~If Y/PY to 2.4: Were these deviations from intended intervention balanced between groups?}

\begin{itemize}[leftmargin=*]
    \item \textbf{NA}: If not Y/PY to 2.4.
    \item \textbf{Y/PY/N/PN}: If the deviation from the established intervention is unrelated to routine medical practice (2.3 answer ``Y/PY''), and there are differences in deviation between groups, this should be taken seriously.
\end{itemize}

\vspace{0.5em}
\textbf{2.6~Was an appropriate analysis used to estimate the effect of assignment to intervention?}

\begin{itemize}[leftmargin=*]
    \item It is considered reasonable to exclude study participants with missing outcome data when applying Intention-to-Treat (ITT) analysis and Modified Intention-to-Treat (mITT) analysis. Inappropriate analytical methods include 'as treated' analysis and 'per-protocol' analysis. When re-grouping after randomization, eligible study participants should not be excluded, but ineligible participants may be excluded.
    \item If a participant is found after randomization to have been ineligible at baseline because they failed to meet the inclusion criteria or met an exclusion criterion, it is methodologically acceptable to remove them completely from the analysis. This practice implies treating such participants as if they were never randomized and does not introduce bias.
\end{itemize}

\vspace{0.5em}
\textbf{2.7~If N/PN/NI to 2.6: Was there potential for a substantial impact (on the result) of the failure to analyse participants in the group to which they were randomised?}

\begin{itemize}[leftmargin=*]
    \item This question primarily focuses on whether the failure to analyze according to the prespecified randomized grouping or the number of study participants not included in the analysis is sufficient to have a significant impact on the results. There is no clear definition of the specific number: when the outcome is a rare event or misclassification is associated with prognostic factors, even if fewer than 5\% of study participants are analyzed in the incorrect group, it may still have an impact on the outcome.
\end{itemize}

\end{tcolorbox}

\vspace{1em}
\begin{tcolorbox}[
    width=\linewidth,
    colback=backblue,
    colframe=gray,
    title={\textcolor{white}{\textbf{Missing Outcome Data Questions}}},
    coltitle=white,
    fonttitle=\bfseries,
    colbacktitle=gray,
    enhanced,
    breakable,
    fontupper=\normalsize
]

\textbf{3.1~Were data for this outcome available for all, or nearly all, participants randomised?}

\begin{itemize}[leftmargin=*]
    \item Intention-to-Treat (ITT) analysis is applicable to situations where all randomized study participants can be included in the analysis. Note: Imputed data are also classified as missing data and are not considered 'outcome data' in this item.
    \item \textbf{NI}: No mention is made in the original text of the extent of missing outcome data; such cases generally carry a high risk of bias due to missing outcome data.
    \item 'Nearly all' study participants means that the number of participants with missing outcomes is small enough that their outcome status does not affect the direction of the estimated intervention effect. For continuous variables, if 95\% (or 90\%) of study participants have available outcome data, this can be considered sufficient. If the outcome is a dichotomous variable, this proportion is related to the probability of the outcome event occurring. If the number of study participants with the outcome event is much larger than the number of those with missing outcome data, only a small degree of bias is introduced.
\end{itemize}

\vspace{0.5em}
\textbf{3.2~If N/PN/NI to 3.1: Is there evidence that the result was not biased by missing outcome data?}

\begin{itemize}[leftmargin=*]
    \item \textbf{NA}: If not N/PN/NI to 3.1.
    \item Such evidence includes: (1) the use of appropriate analytical methods to adjust for bias; (2) the results of sensitivity analysis indicate that, in the context of missing outcomes, the difference between the analytical results and the true results is limited to an acceptable small range. However, when imputing outcome variables, neither the 'Last Observation Carried Forward (LOCF)' method nor multiple imputation based solely on the intervention group is an effective method to remedy the bias caused by missing outcomes.
\end{itemize}

\vspace{0.5em}
\textbf{3.3~If N/PN to 3.2: Could missingness in the outcome depend on its true value?}

\begin{itemize}[leftmargin=*]
    \item If loss to follow-up or withdrawal from the trial is due to participants’ health status, the missing outcome variables are highly likely to be associated with the outcome itself. If all missing outcome variables are not associated with the outcome itself, the risk of bias caused by missing data is low (\eg measurement instrument failure, interruption of data collection, etc.).
\end{itemize}

\vspace{0.5em}
\textbf{3.4~If Y/PY/NI to 3.3: Is it likely that missingness in the outcome depended on its true value?}

\begin{itemize}[leftmargin=*]
    \item This question is classified into the following two scenarios: (1) If missing outcomes are potentially associated with the outcome itself, it is classified as 'some concerns'; (2) If missing outcome variables are highly likely to be associated with the outcome itself, it is classified as 'high concerns'.
    \item For 'high concerns', the answer is 'Y' if the following situations occur: (1) The most likely reason for differences in the proportion of missing outcomes between the two groups is that the missing outcome variables are associated with the outcome itself; (2) The reported reasons for missing outcome variables indicate an association with the outcome itself; (3) The reported reasons for missing outcome variables differ between the two groups; (4) The actual circumstances of the study result in missing outcome variables being highly likely to be associated with the outcome itself. For example: The main reason for withdrawal from schizophrenia-related studies is the patients’ subsequent symptoms.
\end{itemize}

\end{tcolorbox}

\vspace{1em}
\begin{tcolorbox}[
    width=\linewidth,
    colback=backblue,
    colframe=gray,
    title={\textcolor{white}{\textbf{Measurement of the Outcome Questions}}},
    coltitle=white,
    fonttitle=\bfseries,
    colbacktitle=gray,
    enhanced,
    breakable,
    fontupper=\normalsize
]

\textbf{4.1~Was the method of measuring the outcome inappropriate?}

\begin{itemize}[leftmargin=*]
    \item This question aims to evaluate whether outcome measurement during data collection is appropriate, rather than assessing the rationality of the selection of outcome indicators. In general, for prespecified outcomes, the answer to this question is 'N' or 'PN'. If the outcome measurement method is inappropriate, the answer is 'Y' or 'PY', for example: (1) The current measurement method cannot reliably measure the intervention effect (\eg the outcome indicator exceeds the detection range of the measurement method); (2) The measurement tool has poor reliability.
\end{itemize}

\vspace{0.5em}
\textbf{4.2~Could measurement or ascertainment of the outcome have differed between intervention groups?}

\begin{itemize}[leftmargin=*]
    \item The measurement approaches adopted for outcome variables in both groups should be comparable, including the use of the same measurement methods and identical measurement thresholds at comparable time points.
    \item Differences in measurements between the two groups can lead to 'diagnostic detection bias' during outcome data collection; if the intervention group has more clinic visits, this can result in a higher likelihood of identifying the occurrence of outcome events in that group.
\end{itemize}

\vspace{0.5em}
\textbf{4.3~If N/PN/NI to 4.1 and 4.2: Were outcome assessors aware of the intervention received by study participants?}

\begin{itemize}[leftmargin=*]
    \item \textbf{NA}: If not N/PN/NI to 4.1 and 4.2.
    \item If blinding to the intervention status is implemented, the answer to this question is 'N'. For studies with participant-reported outcomes, the outcome assessors are the participants themselves.
\end{itemize}

\vspace{0.5em}
\textbf{4.4~If Y/PY/NI to 4.3: Could assessment of the outcome have been influenced by knowledge of intervention received?}

\begin{itemize}[leftmargin=*]
    \item \textbf{NA}: If not Y/PY/NI to 4.3.
    \item Prior knowledge of the intervention can influence participant-reported outcomes (\eg pain intensity) and lead researchers to introduce subjective judgment when reporting outcomes, thereby affecting outcome measures that rely on the subjective judgment of intervention implementers.
    \item However, if the study outcome does not involve subjective judgment, this will not affect outcome assessment. For example, the outcome may be an objectively determined measure such as death or disease onset.
\end{itemize}

\vspace{0.5em}
\textbf{4.5~If Y/PY/NI to 4.4: Is it likely that assessment of the outcome was influenced by knowledge of intervention received?}

\begin{itemize}[leftmargin=*]
    \item \textbf{NA}: If not Y/PY/NI to 4.4.
    \item This question is classified into the following two scenarios: (1) If knowledge of the intervention potentially influences outcome measurement but there is no evidence that it actually did so, it is classified as 'some concerns'; (2) If knowledge of the intervention is highly likely to influence outcome measurement, it is classified as 'high concerns'. When study participants can anticipate the effects of the intervention, outcome measurement is highly likely to be affected, regardless of whether the effects are expected to be beneficial or harmful. For example: participant-reported symptoms in homeopathy, or physical therapists assessing the recovery of physical function.
\end{itemize}

\end{tcolorbox}

\vspace{1em}
\begin{tcolorbox}[
    width=\linewidth,
    colback=backblue,
    colframe=gray,
    title={\textcolor{white}{\textbf{Selection of the Reported Result Questions}}},
    coltitle=white,
    fonttitle=\bfseries,
    colbacktitle=gray,
    enhanced,
    breakable,
    fontupper=\normalsize
]

\textbf{5.1~Were the data that produced this result analysed in accordance with a prespecified analysis plan that was finalised before unblinded outcome data were available for analysis?}

\begin{itemize}[leftmargin=*]
    \item If the prespecified study protocol has been reported in detail, it is possible to compare the planned outcome measurement methods and analyses with the previously reported protocol. To avoid selective reporting of study results, the final study analysis plan must be developed prior to the unblinding of data to analysts. If the analysis plan is modified before unblinding, or if it can be clearly demonstrated that the modification is unrelated to the results (\eg instrument damage rendering continued data collection impossible), there is no risk of bias from selective reporting of results in such cases.
\end{itemize}

\vspace{0.5em}
\textbf{5.2~Multiple eligible outcome measurements (\eg scales, definitions, time points) within the outcome domain?}

\begin{itemize}[leftmargin=*]
    \item To capture a specific type of outcome indicator, multiple measurement methods may be adopted. For example, pain intensity may involve assessments using multiple scales (\eg Visual Analog Scale [VAS] or McGill Pain Questionnaire [MPQ]) or evaluations at multiple time points (\eg 3, 6, and 12 weeks after treatment). If multiple measurements are performed but only one or several results are reported, a high risk of selective reporting bias arises.
    \item \textbf{Y/PY}: Clear evidence (\eg study protocol or statistical analysis plan [SAP]) demonstrates that multiple measurements were conducted for the outcome, but only one or a few of these measurements were comprehensively reported. In such cases, the comprehensively reported results may be considered selectively reported based on the analysis outcomes. The reason for selective reporting may be the desire to present results more favorable for publication or more conducive to the verification of the research hypothesis. For example, when researchers aim to demonstrate the benefit of the trial group or intervention group, they may be more inclined to report results indicating the effectiveness of the intervention group.
    \item \textbf{N/PN}: Clear evidence (\eg study protocol or statistical analysis plan [SAP]) confirms that all outcome-related measurements were implemented in accordance with the prespecified protocol; or there is only one possible measurement method for the outcome (thus precluding selective reporting); or inconsistencies in outcome measurement methods exist across different reports of the same trial, but the researchers have provided an explanation, and such inconsistencies have no impact on the nature of the results.
    \item \textbf{NI}: The analysis plan is unknown or incompletely reported, and there are multiple measurement methods for the outcome indicator.
\end{itemize}

\vspace{0.5em}
\textbf{5.3~Multiple eligible analyses of the data?}

\begin{itemize}[leftmargin=*]
    \item A specific study outcome may correspond to multiple analytical methods. Examples include: covariate-adjusted and unadjusted models; final values vs.\ changes from baseline vs.\ analysis of covariance (ANCOVA); variable transformations; different definitions of outcome components (\eg 'major adverse reactions'); conversion of continuous variables to categorical variables using different cutoff values; different covariate adjustment approaches; different missing data handling methods. Different analytical methods can yield distinct results for a specific outcome. If multiple analytical results are generated but only one or a few are reported, a high risk of selective reporting bias arises.
    \item \textbf{Y/PY}: Clear evidence (\eg study protocol or statistical analysis plan [SAP]) demonstrates that the outcome was analyzed using multiple approaches, but only one or a few of these analytical results were comprehensively reported. In such cases, the comprehensively reported results may be considered selectively reported based on the analytical outcomes. The reason for selective reporting may be the desire to present results more favorable for publication or more conducive to the verification of the research hypothesis. For example, when researchers aim to demonstrate the benefit of the intervention, they may be more inclined to report results indicating the effectiveness of the trial group.
    \item \textbf{N/PN}: Clear evidence (\eg study protocol or statistical analysis plan [SAP]) confirms that all outcome-related results are consistent with the prespecified analytical methods; or there is only one possible analytical method for the outcome (thus precluding selective reporting); or results from different analytical methods in the same trial are inconsistent, but the researchers have provided an explanation, and such inconsistency is unrelated to the nature of the results.
    \item \textbf{NI}: The analysis plan is unknown or incompletely reported, and there are multiple analytical methods for the outcome indicator.
\end{itemize}

\end{tcolorbox}

\begin{tcolorbox}[
    width=\linewidth,
    colback=backblue,
    colframe=gray,
    title={\textcolor{white}{\textbf{Track B Evidential Faithfulness Prompt}}},
    coltitle=white,
    fonttitle=\bfseries,
    colbacktitle=gray,
    enhanced,
    breakable,
    fontupper=\normalsize
]

\textbf{\# Role: Risk of Bias (RoB) Expert}

You are an expert in critical appraisal of clinical trials using the Cochrane Risk of Bias assessment tool.
Your task is to identify evidence sentences that accurately answer the given question and determine the risk of bias judgment.

\vspace{0.5em}
\textbf{\# Bias Type}

\texttt{\{bias\}}

\vspace{0.5em}
\textbf{\# Question}

\texttt{\{question\}}

\vspace{0.5em}
\textbf{\# Context Sentences}

The following is a list of sentences extracted from the study manuscript. Each sentence is numbered starting from 0.

\texttt{\{context\_list\}}

\vspace{0.5em}
\textbf{\# Task}

You must complete \textbf{two tasks}:

\begin{enumerate}[leftmargin=*]
    \item \textbf{Identify Evidence Sentences}: Determine which sentence indices (0-based) from the \texttt{context\_list} provide accurate evidence to answer the question. You may select multiple sentences if they are all relevant. Return the indices as an array of integers.
    \item \textbf{Determine Risk of Bias}: Based on the evidence sentences you identified, determine the risk of bias judgment according to the criteria provided in the question.
\end{enumerate}

\vspace{0.5em}
\textbf{\# Risk of Bias Judgment}

Allowed values for \texttt{risk\_of\_bias}:
\begin{itemize}[leftmargin=*]
    \item \textbf{low} (Low risk of bias)
    \item \textbf{some concerns} (Some concerns)
    \item \textbf{high} (High risk of bias)
\end{itemize}

\vspace{0.5em}
\textbf{\# Return format (JSON Only)}

You must output a strictly valid JSON object. Do not wrap the JSON in markdown code blocks (like \texttt{```json ... ```}). Output raw JSON only.

Example:\\
\texttt{\{\\
\hspace*{1em}"evidence\_indices": [0, 2, 5],\\
\hspace*{1em}"risk\_of\_bias": "low"\\
\}}

\vspace{0.5em}
\textbf{\# Important Notes}

\textbf{1. Evidence Indices}:
\begin{itemize}[leftmargin=*]
    \item The indices must be 0-based (the first sentence is index 0, the second is index 1, etc.)
    \item You may select multiple sentences if they all provide relevant evidence
    \item If no sentences provide relevant evidence, return an empty array: []
    \item Only include indices that directly support answering the question
\end{itemize}

\vspace{0.5em}
\textbf{2. Risk of Bias Judgment}:
\begin{itemize}[leftmargin=*]
    \item Carefully review the criteria in the question for ``low risk'', ``high risk'', and ``unclear risk''
    \item Base your judgment solely on the evidence sentences you identified
    \item Use the exact values: ``low'', ``some concerns'', or ``high''
\end{itemize}

\vspace{0.5em}
\textbf{3. Output Format}:
\begin{itemize}[leftmargin=*]
    \item Must be valid JSON
    \item Do not include any explanatory text outside the JSON
    \item The JSON object must contain exactly two fields: ``evidence\_indices'' (array of integers) and ``risk\_of\_bias'' (string)
\end{itemize}

\end{tcolorbox}

\begin{tcolorbox}[
    width=\linewidth,
    colback=backblue,
    colframe=gray,
    title={\textcolor{white}{\textbf{CRAG Confidence Scoring Prompt}}},
    coltitle=white,
    fonttitle=\bfseries,
    colbacktitle=gray,
    enhanced,
    breakable,
    fontupper=\normalsize
]

\textbf{\# Role: Evidence Confidence Evaluator}

You are an expert evaluator specializing in assessing the relevance and accuracy of evidence sentences for bias assessment tasks.

\vspace{0.5em}
\textbf{\# Bias Type}

\texttt{\{bias\}}

\vspace{0.5em}
\textbf{\# Question}

\texttt{\{question\}}

\vspace{0.5em}
\textbf{\# Candidate Sentence}

The following is a single sentence from the study manuscript that needs to be evaluated:

Sentence Index: \texttt{\{sentence\_index\}}\\
Sentence: \texttt{\{sentence\}}

\vspace{0.5em}
\textbf{\# Task}

Your task is to evaluate whether this sentence provides accurate evidence to answer the question. You must follow an ``Explain First, Then Grade'' approach:

\textbf{1. First, provide an explanation}: Analyze the sentence against the question's detailed criteria. Identify:
\begin{itemize}[leftmargin=*]
    \item Does the sentence address the core constraints mentioned in the question?
    \item Are there any factual conflicts between the sentence and the question requirements?
    \item Does the sentence provide sufficient information to answer the specific question?
\end{itemize}

\textbf{2. Then, assign a confidence grade}: Based on your analysis, classify the sentence into one of three categories:
\begin{itemize}[leftmargin=*]
    \item \textbf{Correct}: The sentence fully covers the core constraints in the question description, and there are no factual conflicts. The sentence can directly answer the question.
    \item \textbf{Incorrect}: The sentence contradicts key conditions in the question, or although semantically related, it cannot answer the specific question.
    \item \textbf{Uncertain/Ambiguous}: The sentence provides partial information, but due to complex constraint conditions, it is unclear whether it satisfies all detailed requirements.
\end{itemize}

\vspace{0.5em}
\textbf{\# Confidence Scoring (Numerical Logic)}

In addition to the categorical grade, you must also provide a relevance score from 0 to 1:
\begin{itemize}[leftmargin=*]
    \item Score $>$ 0.85 $\rightarrow$ Correct
    \item Score $<$ 0.6 $\rightarrow$ Incorrect
    \item 0.6 $\le$ Score $\le$ 0.85 $\rightarrow$ Uncertain/Ambiguous
\end{itemize}

\vspace{0.5em}
\textbf{\# Return Format (JSON Only)}

You must output a strictly valid JSON object. Do not wrap the JSON in markdown code blocks (like \texttt{```json ... ```}). Output raw JSON only.

Example:\\
\texttt{\{\\
\hspace*{1em}"sentence\_index": 0,\\
\hspace*{1em}"explanation": "This sentence states that [specific information]. It directly addresses the question's requirement about [requirement]. There are no conflicts with the constraints mentioned in the question.",\\
\hspace*{1em}"confidence\_score": 0.92,\\
\hspace*{1em}"confidence\_grade": "Correct"\\
\}}

\vspace{0.5em}
\textbf{\# Important Notes}

\textbf{1. Explanation Requirement}:
\begin{itemize}[leftmargin=*]
    \item You MUST provide a detailed explanation before assigning the grade
    \item The explanation should explicitly compare the sentence against the question's constraints
    \item Identify any potential conflicts or gaps
\end{itemize}

\textbf{2. Confidence Grade}:
\begin{itemize}[leftmargin=*]
    \item Must be exactly one of: ``Correct'', ``Incorrect'', ``Uncertain''
    \item The grade should align with the \texttt{confidence\_score} according to the thresholds above
\end{itemize}

\textbf{3. Confidence Score}:
\begin{itemize}[leftmargin=*]
    \item Must be a number between 0 and 1 (inclusive)
    \item Should reflect the degree to which the sentence satisfies the question requirements
\end{itemize}

\textbf{4. Output Format}:
\begin{itemize}[leftmargin=*]
    \item Must be valid JSON
    \item Do not include any explanatory text outside the JSON
    \item The JSON object must contain exactly four fields: ``sentence\_index'' (integer), ``explanation'' (string), ``confidence\_score'' (number), and ``confidence\_grade'' (string)
\end{itemize}

\end{tcolorbox}

\begin{tcolorbox}[
    width=\linewidth,
    colback=backblue,
    colframe=gray,
    title={\textcolor{white}{\textbf{CRAG Reflection Prompt}}},
    coltitle=white,
    fonttitle=\bfseries,
    colbacktitle=gray,
    enhanced,
    breakable,
    fontupper=\normalsize
]

\textbf{\# Role: Evidence Reflection Analyzer}

You are an expert analyzer specializing in deep reflection and verification of uncertain evidence sentences using the Divide-Verify-Refine (DVR) strategy.

\vspace{0.5em}
\textbf{\# Bias Type}

\texttt{\{bias\}}

\vspace{0.5em}
\textbf{\# Question}

\texttt{\{question\}}

\vspace{0.5em}
\textbf{\# Uncertain Sentence}

The following sentence was initially classified as ``Uncertain/Ambiguous'' and requires deeper analysis:

Sentence Index: \texttt{\{sentence\_index\}}\\
Sentence: \texttt{\{sentence\}}

\vspace{0.5em}
\textbf{\# Task}

You must perform a three-step DVR (Divide-Verify-Refine) reflection process:

\vspace{0.5em}
\textbf{\#\# Step 1: Divide (Decompose)}

Break down the complex instructions in the question description into independent atomic constraints. Identify:
\begin{itemize}[leftmargin=*]
    \item Temporal requirements (\eg specific time periods, dates)
    \item Geographic restrictions (\eg location-specific conditions)
    \item Specific numerical values or thresholds
    \item Methodological requirements
    \item Outcome-related constraints
    \item Any other explicit or implicit constraints
\end{itemize}
List each atomic constraint clearly.

\vspace{0.5em}
\textbf{\#\# Step 2: Verify (Create Verification Matrix)}

For each atomic constraint identified in Step 1, verify whether the uncertain sentence satisfies it. Create a verification matrix with the following structure:

For each constraint:
\begin{itemize}[leftmargin=*]
    \item Constraint Description: [What the constraint requires]
    \item Sentence Mentions: [Yes/No/Unknown]
    \item Evidence in Sentence: [Specific text or information from the sentence that relates to this constraint]
    \item Satisfies Constraint: [Yes/No/Partially/Unknown]
    \item Reasoning: [Brief explanation of why this constraint is or is not satisfied]
\end{itemize}

\vspace{0.5em}
\textbf{\#\# Step 3: Refine (Re-evaluate)}

Based on the verification matrix from Step 2, re-evaluate the sentence's confidence grade:
\begin{itemize}[leftmargin=*]
    \item If ALL core constraints are satisfied $\rightarrow$ Upgrade to ``Correct''
    \item If there are explicit conflicts with core constraints $\rightarrow$ Downgrade to ``Incorrect''
    \item If some constraints are satisfied but others remain unclear $\rightarrow$ Keep as ``Uncertain'' but provide detailed reasoning
\end{itemize}

\vspace{0.5em}
\textbf{\# Return Format (JSON Only)}

You must output a strictly valid JSON object. Do not wrap the JSON in markdown code blocks (like \texttt{```json ... ```}). Output raw JSON only.

Example:\\
\texttt{\{\\
\hspace*{1em}"sentence\_index": 0,\\
\hspace*{1em}"divide": \{\\
\hspace*{2em}"atomic\_constraints": [\\
\hspace*{3em}"Constraint 1: [description]",\\
\hspace*{3em}"Constraint 2: [description]",\\
\hspace*{3em}"Constraint 3: [description]"\\
\hspace*{2em}]\\
\hspace*{1em}\},\\
\hspace*{1em}"verify": \{\\
\hspace*{2em}"verification\_matrix": [\\
\hspace*{3em}\{\\
\hspace*{4em}"constraint\_description": "Constraint 1: [description]",\\
\hspace*{4em}"sentence\_mentions": "Yes",\\
\hspace*{4em}"evidence\_in\_sentence": "[specific text]",\\
\hspace*{4em}"satisfies\_constraint": "Yes",\\
\hspace*{4em}"reasoning": "[explanation]"\\
\hspace*{3em}\},\\
\hspace*{3em}\{\\
\hspace*{4em}"constraint\_description": "Constraint 2: [description]",\\
\hspace*{4em}"sentence\_mentions": "No",\\
\hspace*{4em}"evidence\_in\_sentence": "N/A",\\
\hspace*{4em}"satisfies\_constraint": "No",\\
\hspace*{4em}"reasoning": "[explanation]"\\
\hspace*{3em}\}\\
\hspace*{2em}]\\
\hspace*{1em}\},\\
\hspace*{1em}"refine": \{\\
\hspace*{2em}"final\_confidence\_grade": "Correct",\\
\hspace*{2em}"confidence\_score": 0.88,\\
\hspace*{2em}"reasoning": "After verification, all core constraints are satisfied. The sentence mentions [key information] which directly addresses [constraint]. Therefore, the sentence is upgraded to Correct."\\
\hspace*{1em}\}\\
\}}

\vspace{0.5em}
\textbf{\# Important Notes}

\textbf{1. Divide Step}:
\begin{itemize}[leftmargin=*]
    \item Be thorough in identifying ALL constraints from the question
    \item Each constraint should be atomic and independently verifiable
    \item Include both explicit and implicit constraints
\end{itemize}

\textbf{2. Verify Step}:
\begin{itemize}[leftmargin=*]
    \item The verification matrix must cover ALL constraints identified in Step 1
    \item Be specific about what evidence exists (or doesn't exist) in the sentence
    \item Provide clear reasoning for each constraint verification
\end{itemize}

\textbf{3. Refine Step}:
\begin{itemize}[leftmargin=*]
    \item The final \texttt{confidence\_grade} must be one of: ``Correct'', ``Incorrect'', ``Uncertain''
    \item The \texttt{confidence\_score} should reflect the refined assessment (0--1 scale)
    \item Provide detailed reasoning explaining the upgrade/downgrade decision
\end{itemize}

\textbf{4. Output Format}:
\begin{itemize}[leftmargin=*]
    \item Must be valid JSON
    \item Do not include any explanatory text outside the JSON
    \item The JSON structure must include: ``sentence\_index'', ``divide'', ``verify'', and ``refine'' fields
\end{itemize}

\end{tcolorbox}

\begin{tcolorbox}[
    width=\linewidth,
    colback=backblue,
    colframe=gray,
    title={\textcolor{white}{\textbf{Chain of Thought (CoT) Prompt}}},
    coltitle=white,
    fonttitle=\bfseries,
    colbacktitle=gray,
    enhanced,
    breakable,
    fontupper=\normalsize
]

\textbf{\# Role: Risk of Bias (RoB) Expert}

You are an expert in critical appraisal of clinical trials using the Cochrane Risk of Bias assessment tool.
Your task is to identify evidence sentences that accurately answer the given question and determine the risk of bias judgment.

\vspace{0.5em}
\textbf{\# Bias Type}

\texttt{\{bias\}}

\vspace{0.5em}
\textbf{\# Question}

\texttt{\{question\}}

\vspace{0.5em}
\textbf{\# Context Sentences}

The following is a list of sentences extracted from the study manuscript. Each sentence is numbered starting from 0.

\texttt{\{context\_list\}}

\vspace{0.5em}
\textbf{\# Chain of Thought (CoT) Reasoning}

\textbf{IMPORTANT: You must use Chain of Thought reasoning for both tasks.}

\textbf{For identifying evidence sentences}:

\begin{enumerate}[leftmargin=*]
    \item \textbf{Understand the question}:
    \begin{itemize}
        \item What specific information is the question asking for?
        \item What are the key criteria mentioned in the question?
        \item What would constitute evidence for ``low risk'', ``high risk'', or ``some concerns''?
    \end{itemize}
    \item \textbf{Review each sentence systematically}:
    \begin{itemize}
        \item Read each sentence in the context list carefully
        \item For each sentence, ask: ``Does this sentence provide information relevant to the question?''
        \item Consider: ``Does this sentence directly address the criteria mentioned in the question?''
    \end{itemize}
    \item \textbf{Evaluate relevance and accuracy}:
    \begin{itemize}
        \item Identify sentences that directly answer the question
        \item Distinguish between sentences that are relevant vs.\ those that are not
        \item Consider whether multiple sentences together provide a complete answer
    \end{itemize}
    \item \textbf{Select evidence sentences}:
    \begin{itemize}
        \item Choose sentences that provide the most direct and accurate evidence
        \item Include all sentences that are necessary to fully answer the question
        \item Exclude sentences that are irrelevant or misleading
    \end{itemize}
\end{enumerate}

\textbf{For determining risk of bias}:

\begin{enumerate}[leftmargin=*]
    \item \textbf{Review the selected evidence}:
    \begin{itemize}
        \item What information do the evidence sentences provide?
        \item How does this information relate to the question criteria?
    \end{itemize}
    \item \textbf{Apply the criteria from the question}:
    \begin{itemize}
        \item What are the specific criteria for ``low risk of bias''?
        \item What are the specific criteria for ``high risk of bias''?
        \item What are the specific criteria for ``some concerns''?
    \end{itemize}
    \item \textbf{Match evidence to criteria}:
    \begin{itemize}
        \item Does the evidence meet the criteria for ``low risk''? If so, how?
        \item Does the evidence indicate ``high risk''? If so, what specific concerns?
        \item Is there uncertainty that leads to ``some concerns''?
    \end{itemize}
    \item \textbf{Make the judgment}:
    \begin{itemize}
        \item Based on your analysis, which risk level best matches the evidence?
        \item Justify your choice by explaining how the evidence aligns with the criteria
    \end{itemize}
\end{enumerate}

\vspace{0.5em}
\textbf{\# Task}

You must complete \textbf{two tasks}:

\begin{enumerate}[leftmargin=*]
    \item \textbf{Identify Evidence Sentences}: Determine which sentence indices (0-based) from the \texttt{context\_list} provide accurate evidence to answer the question. You may select multiple sentences if they are all relevant. Return the indices as an array of integers.
    \item \textbf{Determine Risk of Bias}: Based on the evidence sentences you identified, determine the risk of bias judgment according to the criteria provided in the question.
\end{enumerate}

\vspace{0.5em}
\textbf{\# Risk of Bias Judgment}

Allowed values for \texttt{risk\_of\_bias}:
\begin{itemize}[leftmargin=*]
    \item \textbf{low} (Low risk of bias)
    \item \textbf{some concerns} (Some concerns)
    \item \textbf{high} (High risk of bias)
\end{itemize}

\vspace{0.5em}
\textbf{\# Return format (JSON Only)}

You must output a strictly valid JSON object. Do not wrap the JSON in markdown code blocks (like \texttt{```json ... ```}). Output raw JSON only.

Example:\\
\texttt{\{\\
\hspace*{1em}"evidence\_indices": [0, 2, 5],\\
\hspace*{1em}"risk\_of\_bias": "low",\\
\hspace*{1em}"reasoning": "Step-by-step explanation: First, I analyzed the question which asks about [topic]. I reviewed all sentences and identified that sentences 0, 2, and 5 directly address this question. Sentence 0 states [evidence], sentence 2 provides [evidence], and sentence 5 indicates [evidence]. According to the criteria, [analysis]. Therefore, I selected [risk\_of\_bias] because [justification]."\\
\}}

\vspace{0.5em}
\textbf{\# Important Notes}

\textbf{1. Evidence Indices}:
\begin{itemize}[leftmargin=*]
    \item The indices must be 0-based (the first sentence is index 0, the second is index 1, etc.)
    \item You may select multiple sentences if they all provide relevant evidence
    \item If no sentences provide relevant evidence, return an empty array: []
    \item Only include indices that directly support answering the question
\end{itemize}

\textbf{2. Risk of Bias Judgment}:
\begin{itemize}[leftmargin=*]
    \item Carefully review the criteria in the question for ``low risk'', ``high risk'', and ``unclear risk''
    \item Base your judgment solely on the evidence sentences you identified
    \item Use the exact values: ``low'', ``some concerns'', or ``high''
\end{itemize}

\textbf{3. Chain of Thought Reasoning}:
\begin{itemize}[leftmargin=*]
    \item The \texttt{reasoning} field is required and must demonstrate your step-by-step thinking process
    \item Explain how you identified the evidence sentences
    \item Show how you evaluated the evidence against the criteria
    \item Justify your risk of bias judgment
\end{itemize}

\textbf{4. Output Format}:
\begin{itemize}[leftmargin=*]
    \item Must be valid JSON
    \item Do not include any explanatory text outside the JSON
    \item The JSON object must contain exactly three fields: ``evidence\_indices'' (array of integers), ``risk\_of\_bias'' (string), and ``reasoning'' (string)
\end{itemize}

\end{tcolorbox}

\end{document}